\documentstyle[chum]{article}

\title{Japanese/English Cross-Language Information Retrieval:
Exploration of Query Translation and Transliteration\footnote{Computers and the Humanities, Vol.35, No.4, pp.389--420, Nov. 2001}}

\author{\Large Atsushi Fujii and Tetsuya Ishikawa}

\date{University of Library and Information Science \\
1-2 Kasuga Tsukuba 305-8550, JAPAN \\ \smallskip
{\normalsize\tt 
E-mail:fujii@ulis.ac.jp}}

\summary{Cross-language information retrieval (CLIR), where queries
and documents are in different languages, has of late become one of
the major topics within the information retrieval community. This
paper proposes a Japanese/English CLIR system, where we combine a
query translation and retrieval modules. We currently target the
retrieval of technical documents, and therefore the performance of our
system is highly dependent on the quality of the translation of
technical terms.  However, the technical term translation is still
problematic in that technical terms are often compound words, and thus
new terms are progressively created by combining existing base
words. In addition, Japanese often represents loanwords based on its
special phonogram. Consequently, existing dictionaries find it
difficult to achieve sufficient coverage. To counter the first
problem, we produce a Japanese/English dictionary for base words, and
translate compound words on a word-by-word basis.  We also use a
probabilistic method to resolve translation ambiguity.  For the second
problem, we use a transliteration method, which corresponds words
unlisted in the base word dictionary to their phonetic equivalents in
the target language. We evaluate our system using a test collection
for CLIR, and show that both the compound word translation and
transliteration methods improve the system performance.}

\begin{document}

\makeidpage
\maketitle

\input{psfig.tex}

\newcommand{\etal}{et~al.}
\newcommand{\etaleos}{et~al}
\newcommand{\eq}[1]{(\ref{#1})}

\renewcommand{\nocite}[1]{\shortcite{#1}}

\section{Introduction}
\label{sec:introduction}

Cross-language information retrieval (CLIR) is the retrieval process
where the user presents queries in one language to retrieve documents
in {\em another\/} language. One of the traditional research
references for CLIR dates back to the 1960s~\cite{mongar:tis-69}. In
the 1970s, Salton~\nocite{salton:jasis-70,salton:techrep-72}
empirically showed that CLIR using a hand-crafted bilingual thesaurus
is comparable with monolingual information retrieval in
performance. The 1990s witnessed a growing number of machine readable
texts in various languages, including those accessible via the World
Wide Web, but each content is usually provided in a limited number of
languages. Thus, it is feasible that users are interested in
retrieving information across languages. Possible users of CLIR are
given below:
\begin{itemize}
\item users who are able to read documents in foreign languages, but
  have difficulty formulating foreign queries,
\item users who find it difficult to retrieve/read relevant documents,
  but need the information, for the purpose of which the use of
  machine translation (MT) systems for the limited number of documents
  retrieved through CLIR is computationally more efficient rather than
  translating the entire collection,
\item users who know foreign keywords/phrases, and want to read
  documents associated with them, in their native language.
\end{itemize}
In fact, CLIR has of late become one of the major topics within the
information retrieval (IR), natural language processing (NLP) and
artificial intelligence (AI) communities, and numerous CLIR systems
have variously been
proposed~\cite{aaai-spring-sympo-97,sigir-96-98,trec-92-98}.
Note that CLIR can be seen as a subtask of multi-lingual information
retrieval (MLIR), which also includes the following cases:
\begin{itemize}
\item identify the query language (based on, for example, character
  codes), and search a multilingual collection for documents in the
  query language,
\item retrieve documents, in which each document is in more than one
  language,
\item retrieve documents using a query in more than one
  language~\cite{fung:acl-99}.
\end{itemize}
However, these above cases are beyond the scope of this paper. It
should also be noted that while CLIR is not necessarily limited to IR
within two languages, we consistently use the term ``bilingual,''
keeping the potential applicability of CLIR to more than two languages
in mind, because the variety of languages used is not the central
issue of this paper.

Since by definition queries and documents are in different languages,
CLIR needs a translation process along with the conventional
monolingual retrieval process.  For this purpose, existing CLIR
systems adopt various techniques explored in NLP research. In brief,
dictionaries, corpora, thesauri and MT systems are used to translate
queries and/or documents. However, due to the rudimentary nature of
existing translation methods, CLIR still finds it difficult to achieve
the performance of monolingual IR. Roughly speaking, recent
experiments showed that the average precision of CLIR is 50-75\% of
that obtained with monolingual IR~\cite{schauble:trec-97}, which
stimulates us to further explore this exciting research area.

In this paper, we propose a Japanese/English bidirectional CLIR system
targeting technical documents, which has been less explored than that
for newspaper articles in past CLIR literature.  Our research is
partly motivated by the NACSIS test collection for (CL)IR systems,
which consists of Japanese queries and Japanese/English abstracts
collected from technical papers~\cite{kando:sigir-99}.\footnote{\tt
{http://www.rd.nacsis.ac.jp/\~{}ntcadm/index-en.html}} We will
elaborate on the NACSIS collection in
Section~\ref{subsec:eval_overview}. As can be predicted, the
performance of our CLIR system strongly depends on the quality of the
translation of technical terms, which are often unlisted in general
dictionaries.

Pirkola~\nocite{pirkola:sigir-98}, for example, used a subset of the
TREC collection related to health topics, and showed that a
combination of general and domain specific (i.e., medical)
dictionaries improves the CLIR performance obtained with only a
general dictionary. This result shows the potential contribution of
technical term translation to CLIR. At the same time, it should be
noted that even domain specific dictionaries do not exhaustively list
possible technical terms. For example, the EDR technical terminology
dictionary~\cite{edr-techdic:95}, which consists of approximately
120,000 Japanese-English translations related to the information
processing field, lacks recent terms like ``{\it jouhou
chuushutsu\/}~(information extraction).'' We classify problems
associated with technical term translation as given below:
\begin{itemize}
\item technical terms are often compound words, which
  can be progressively created simply by combining multiple existing
  morphemes (``base words''), and therefore it is not entirely
  satisfactory or feasible to exhaustively enumerate newly emerging
  terms in dictionaries,
\item Japanese often represents loanwords (i.e., technical terms and
  proper nouns imported from foreign languages) using its special
  phonetic alphabet (or phonogram) called ``{\it katakana},'' with
  which new words can be spelled out,
\item English technical terms are often abbreviated, which can be used
  as ``Japanese'' words.
\end{itemize}
To counter the first problem, we propose a compound word translation
method, which selects appropriate translations based on the
probability of occurrence of each combination of base words in the
target language (see Section~\ref{subsec:cwt}). Note that technical
compound words sometimes include general words, such as ``AI {\em
chess\/}'' and ``digital {\em watermark\/}.'' In this paper, we do not
rigorously define general words, by which we mean words that are
contained in existing general dictionaries but rarely in technical
term dictionaries. For the second problem, we propose a
``transliteration'' method, which identifies phonetic equivalents in
the target language (see Section~\ref{subsec:translit}). Finally, to
resolve the third problem, we enhance our bilingual dictionary with
multiples of each abbreviation and its complete form (e.g., ``IR'' and
``information retrieval'') extracted from English corpora (see
Section~\ref{subsec:dictionary_enhancement}).  Note that although a
number of methods targeting those above problems have been explored
in past research, no attempt has been made to integrate them in the
context of CLIR.

Section~\ref{sec:past_research} surveys past research on CLIR, and
clarifies our focus and approach. Section~\ref{sec:system_overview}
overviews our CLIR system, and Section~\ref{sec:translation}
elaborates on the translation method aimed to resolve the above
problems associated with technical term translation.
Section~\ref{sec:evaluation} then evaluates the performance of our
CLIR system using the NACSIS collection.

\section{Past Research on CLIR}
\label{sec:past_research}

\subsection{Retrieval Methodologies}
\label{subsec:retrieval_methods}

\begin{figure*}[htbp]
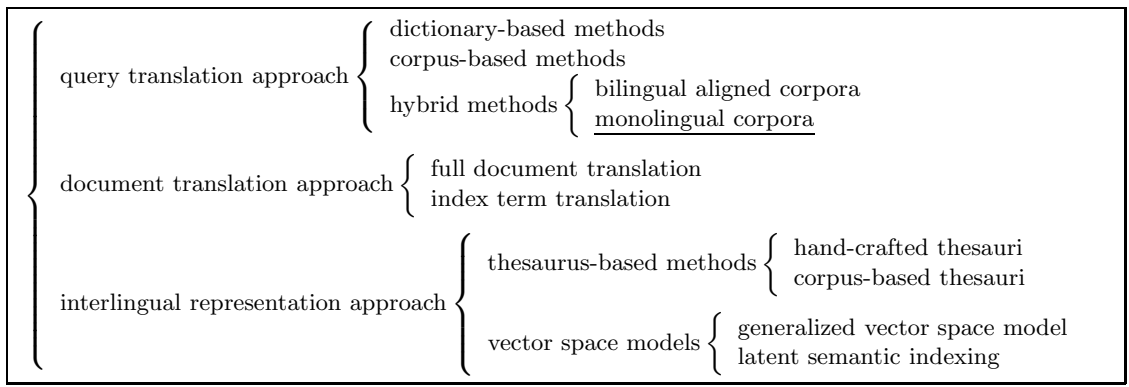

  \def\baselinestretch{1}
  \begin{center}
    \leavevmode
    \small
    \fbox{
    $\left\{
    \begin{array}{l}
      \mbox{query translation approach}\left\{
        \begin{array}{l}
          \mbox{dictionary-based methods} \\
          \mbox{corpus-based methods} \\
          \mbox{hybrid methods}\left\{
          \begin{array}{l}
            \mbox{bilingual aligned corpora} \\
            $\underline{\mbox{monolingual corpora}}$
          \end{array}\right.
        \end{array}\right. \medskip \\
      \mbox{document translation approach}\left\{
        \begin{array}{l}
          \mbox{full document translation} \\
          \mbox{index term translation}
        \end{array}\right. \medskip \\
      \mbox{interlingual representation approach}\left\{
        \begin{array}{l}
          \mbox{thesaurus-based methods}\left\{
            \begin{array}{l}
              \mbox{hand-crafted thesauri} \\
              \mbox{corpus-based thesauri}
            \end{array}\right.\medskip \\
          \mbox{vector space models}\left\{
            \begin{array}{l}
              \mbox{generalized vector space model} \\
              \mbox{latent semantic indexing}
            \end{array}\right.
        \end{array}\right.
    \end{array}
    \right.$}
  \end{center}
  \medskip
  \caption{Classification of CLIR retrieval methods (the method we
  adopt is underlined)}
  \label{fig:retrieval_methods}
\end{figure*}

Figure~\ref{fig:retrieval_methods} classifies existing CLIR
approaches in terms of retrieval methodology. The top level three
categories correspond to the different titles of the following items.

\paragraph{Query translation approach}

This approach translates queries into document languages using
bilingual dictionaries or/and corpora, prior to the retrieval process.
Since the retrieval process is fundamentally the same as performed in
monolingual IR, the translation module can easily be combined with
existing IR engines. This category can be further subdivided into the
following three methods.

The first subcategory can be called dictionary-based methods.  Hull
and Grefenstette~\nocite{hull:sigir-96} used a bilingual dictionary to
derive all possible translation candidates of query terms, which are
used for the subsequent retrieval. Their method is easy to implement,
but potentially retrieves irrelevant documents and decreases the time
efficiency. To resolve this problem,
Hull~\nocite{hull:aaai-spring-sympo-97} combined translation
candidates for each query term with the ``OR'' operator, and used the
weighted boolean method to assign an importance degree to each
translation candidate.

Pirkola~\nocite{pirkola:sigir-98} also used structured queries, where
each term is combined with different types of operators. Ballesteros
and Croft~\nocite{ballesteros:sigir-97} enhanced the dictionary-based
translation using the ``local context analysis''~\cite{xu:sigir-96}
and phrase-based translation.  Dorr and Oard~\nocite{dorr:lrec-98}
evaluated the effectiveness of a semantic structure of a query in the
query translation. As far as their comparative experiments were
concerned, the use of semantic structures was not as effective as
MT/dictionary-based query translation methods.

The second subcategory, corpus-based methods, uses translations
extracted from bilingual corpora, for the query
translation~\cite{carbonell:ijcai-97}. In this paper, ``(bilingual)
aligned corpora'' generally refer to a pair of two language corpora
aligned to each other on a word, sentence, paragraph or document
basis. Given such resources, corpus-based methods are expected to
acquire domain specific translations unlisted in existing
dictionaries. In fact, Carbonell~\etal~\nocite{carbonell:ijcai-97}
empirically showed that their corpus-based query translation method
outperformed a dictionary-based method.  Their comparative evaluation
also showed that the corpus-based translation method outperformed
GVSM/LSI-based methods (see the following ``Interlingual
representation approach'' item for details of GVSM and LSI). Note that
for the purpose of corpus-based translation methods, a number of
translation extraction techniques explored in NLP
research~\cite{fung:acl-95,kaji:coling-96,smadja:cl-96} are
applicable.

Finally, hybrid methods use corpora to resolve the translation
ambiguity inherent in bilingual dictionaries.  Unlike the corpus-based
translation methods described above, which rely on bilingual corpora,
Ballesteros and Croft~\nocite{ballesteros:sigir-98} and
Chen~\etal~\nocite{chen:acl-99} independently used a {\em
monolingual\/} corpus for the disambiguation, and therefore the
implementation cost is less. In practice, their method selects the
combination of translation candidates that frequently co-occur in the
target language corpus. On the other hand, bilingual corpora are also
applicable to hybrid
methods. Okumura~\etal~\nocite{okumura:lrec-tlim-ws-98} and
Yamabana~\etal~\nocite{yamabana:sigir-ws-96} independently used the
same disambiguation method, in that they consider word frequencies in
both the source and target languages, obtained from a bilingual
aligned corpus. Nie~\etal~\nocite{nie:sigir-99} automatically
collected parallel texts in French and English from the World Wide
Web, to train a probabilistic query translation model, and suggested
its feasibility for CLIR.

Davis and Ogden~\nocite{davis:sigir-97} used a bilingual aligned
corpus as the document collection for training retrieval. They first
derive possible translation candidates using a dictionary. Then,
training retrieval trials are performed on the bilingual corpus, in
which the source and translated queries are used to retrieve source
and target documents, respectively. Finally, they select translations
which retrieved documents aligned to those retrieved with the source
query. Note that this method provides a salient contrast to other
query translation methods, in which translation is performed
independently from the retrieval
module.

Chen~\etal~\nocite{chen:acl-99} addressed the disambiguation of
polysemy in the target language, along with the translation
disambiguation, specifically in the case where a source query term
corresponds to a small number of translations, but some of these
translations are associated with a large number of word senses, the
polysemous disambiguation is more crucial than the resolution of
translation ambiguity. To counter this problem, source query terms are
expanded with words that frequently co-occur, which are expected to
restrict the meaning of polysemous words in the target language
documents.

\paragraph{Document translation approach}

This approach translates documents into query languages, prior to the
retrieval. In most cases, existing MT systems are used to translate
all the documents in a given
collection~\cite{gachot:sigir-ws-96,kwon:cpol-98,oard:amta-98}. Otherwise,
a dictionary-based method is used to translate only index
terms~\cite{aone:anlp-97}. It is feasible that when compared with
short queries, documents contain a significantly higher volume of
information for the translation. In fact, Oard~\nocite{oard:amta-98}
showed that the document translation method using an MT system
outperformed several types of dictionary-based query translation
methods.

However, McCarley~\nocite{mccarley:acl-99} showed that the relative
superiority between query and document translation approaches varied
depending on the source and target language pair. He also showed that
a hybrid system (it should not be confused with one described in the
``Query translation approach'' item above), where the relevance degree
of each document (i.e., the ``score'') is the mean of those obtained
with query and document translation systems, outperformed systems
based on either query or document translation approach. However,
generally speaking, the full translation on large-scale collections
can be prohibitive.

\paragraph{Interlingual representation approach}

The basis of this approach is to project both queries and documents in
a language-independent (conceptual) space. In other words, as
Salton~\nocite{salton:jasis-70,salton:techrep-72} and Sheridan and
Ballerini~\nocite{sheridan:sigir-96} identified, the interlingual
representation approach is based on query expansion methods proposed
for monolingual IR. This category can be subdivided into
thesaurus-based methods and variants of the vector space model
(VSM)~\cite{salton:83}.

Salton~\nocite{salton:jasis-70,salton:techrep-72} applied hand-crafted
English/French and English/German thesauri to the SMART
system~\cite{salton:71}, and demonstrated that a CLIR version of the
SMART system is comparable to the monolingual version in
performance. The International Road Research Documentation
scheme~\cite{mongar:tis-69} used a trilingual thesaurus associated
with English, German and French.
Gilarranz~\etal~\nocite{gilarranz:aaai-spring-sympo-97} and
Gonzalo~\etal~\nocite{gonzalo:chum-98} used the EuroWordNet
multilingual thesaurus~\cite{vossen:chum-98}.  Unlike these above
methods relying on manual thesaurus construction, Sheridan and
Ballerini~\nocite{sheridan:sigir-96} used a multilingual thesaurus
automatically produced from an aligned corpus.

The generalized vector space model (GVSM)~\cite{wong:sigir-85} and
latent semantic indexing (LSI)~\cite{deerwester:jasis-90}, which were
originally proposed as variants of the vector space model for
monolingual IR, project both queries and documents into a
language-independent vector space, and therefore these methods can be
applicable to CLIR. While Dumais~\etal~\nocite{dumais:sigir-ws-96}
explored an LSI-based CLIR,
Carbonell~\etal~\nocite{carbonell:ijcai-97} empirically showed that
GVSM outperformed LSI in terms of CLIR. Note that like thesaurus-based
methods, GVSM/LSI-based methods require aligned corpora.

\subsection{Presentation Methodologies}
\label{subsec:presentation_methods}

In the case of CLIR, retrieved documents are not always written in the
user's native language. Therefore, presentation methodology of
retrieval results is a more crucial task than in monolingual IR.  It
is desirable to present smaller-sized contents with less noise, in
other words, precision is often given more importance than recall for
CLIR systems. Note that effective presentation is also crucial when a
user and system interactively retrieve relevant documents, as
performed in relevance feedback~\cite{salton:83}.

However, a surprisingly small number of references addressing this
issue can be found in past research literature.
Aone~\etal~\nocite{aone:anlp-97} presented only keywords frequently
appearing in retrieved documents, rather than entire documents. Note
that since most CLIR systems use frequency information associated with
index terms like ``term frequency (TF)'' and ``inverse document
frequency (IDF)''~\cite{salton:83} for the retrieval, frequently
appearing keywords can be identified without an excessive additional
computational cost. Experiments independently conducted by Oard and
Resnik~\nocite{oard:ipm-99} and
Suzuki~\etal~\nocite{suzuki:signl-98-7} showed that even a simple
translation of keywords (such as using all possible translations
defined in a dictionary) improved on the efficiency for users to find
relevant foreign documents from the whole retrieval result.
Suzuki~\etal~\nocite{suzuki:nlp-99} more extensively investigated the
user's retrieval efficiency (i.e., the time efficiency and accuracy
with which human subjects find relevant foreign documents) by
comparing different presentation methods, in which the following
contents were independently presented to the user:
\begin{enumerate}
\item keywords without translation,
\item keywords translated with the first entry defined in a dictionary,
\item keywords translated through the hybrid method (see the
  ``Query translation approach'' item in
  Section~\ref{subsec:retrieval_methods}),
\item documents summarized (by an existing summarization software) and
  manually translated.
\end{enumerate}
Their comparative experiments showed that the third content was most
effective in terms of the retrieval efficiency.

For monolingual IR, automatic summarization methods based on the
user's focus/query have recently been explored. Mani and
Bloedorn~\nocite{mani:aaai-iaai-98} used machine learning techniques
to produce document summarization rules based on the user's focus
(i.e., query). Tombros and Sanderson~\nocite{tombros:sigir-98} showed
experimental results, in which presenting the fragment of each
retrieved document containing query terms improved on the retrieval
efficiency of human subjects. Applicability of these methods to CLIR
needs to be further explored.

\subsection{Evaluation Methodologies}
\label{subsec:evaluation_methods}

From a scientific point of view, performance evaluation is invaluable
for CLIR. In most cases, the evaluation of CLIR is the same as
performed for monolingual IR. That is, each system conducts a
retrieval trial using a test collection consisting of predefined
queries and documents in {\em different\/} languages, and then the
performance is evaluated based on the precision and recall. Several
experiments used test collections for monolingual IR in which either
queries or documents were translated, prior to the
evaluation. However, as Sakai~\etal~\nocite{sakai:tipsj-99}
empirically showed, the CLIR performance varies depending on the
quality of the translation of collections, and thus it is desirable to
carefully produce test collections for CLIR. The production of test
collections usually involves collecting documents, producing queries
and relevance assessment for each query. However, since relevance
assessment is expensive, especially for large-scale collections (even
in the case where the pooling method~\cite{voorhees:sigir-98} is used
to reduce the number of candidates of relevant documents),
Carbonell~\etal~\nocite{carbonell:ijcai-97} first translated queries
into the document language, and used as (pseudo) relevant documents
those retrieved with the translated queries. In other words, this
evaluation method investigates the extent to which CLIR maintains the
performance of monolingual IR.

For the evaluation of presentation methods, human subjects are often
used to investigate the retrieval efficiency, as described in
Section~\ref{subsec:presentation_methods}. However, evaluation methods
involving human interactions are problematic, because human subjects
are in a way trained through repetitive retrieval trials for different
systems, which can potentially bias the result. On the other hand, in
the case where each subject uses a single system, difference of
subjects affects the result. To minimize this bias, multiple subjects
are usually classified based on, for example, their literacy in terms
of the target language, and those falling into the same cluster are
virtually regarded as the same person.  However, this issue still
remains an open question, and needs to be further explored.

\subsection{Our Focus and Approach}
\label{subsec:our_approach}

Through discussions in the above three sections, we identified the
following points which should be taken into consideration for our
research.

For translation methodology, the query translation approach is
preferable in terms of implementation cost, because this approach can
simply be combined with existing IR engines. On the other hand, other
approaches can be prohibitive, because (a) the document translation
approach conducts the full translation on the entire collection, and
(b) the interlingual representation approach requires alignment of
bilingual thesauri/corpora. In fact, we do not have Japanese-English
thesauri/corpora with sufficient volume of alignment information at
present. One may argue that the NACSIS collection, which is a
large-scale Japanese-English aligned corpora, can be used for the
translation. However, note that bilingual corpora for the translation
must not be obtained from the test collection used for the evaluation,
because in real world usage one of the two language documents in the
collection is usually missing. In other words, CLIR has little
necessity for bilingual aligned document collections, in that the user
can retrieve documents in the query language, without the translation
process.

However, at the same time we concede that each approach is worth
further exploration, and in this paper we do not pretend to draw any
premature conclusions regarding the relative merits of different
approaches. To sum up, we focus mainly on translating sequences of
content words included in queries, rather than the entire
collection. Among different methods following the query translation
approach, we adopt the hybrid method using a {\em monolingual\/}
corpus. In other words, our translation method is relatively similar
to that proposed by Ballesteros and
Croft~\etal~\nocite{ballesteros:sigir-98} and
Chen~\etal~\nocite{chen:acl-99}. However, unlike their cases, we
integrate word-based translation and transliteration methods within
the query translation.

For presentation methodology, we use keywords translated using the
hybrid translation method, which were proven to be effective in
comparative experiments by Suzuki~\etal~\nocite{suzuki:nlp-99} (in the
case where retrieved documents are not in the user's native language).
Note that for the purpose of the translation of keywords, we can use
exactly the same method as performed for the query translation,
because both queries and keywords usually consist of one or more
content words.

Finally, for the evaluation of our CLIR system we use the NACSIS
collection~\cite{kando:sigir-99}. Since in this collection relevance
assessment is performed between Japanese queries and Japanese/English
documents, we can easily evaluate our system in terms of
Japanese-English CLIR. On the other hand, the evaluation of
English-Japanese CLIR is beyond the scope of this paper, because as
discussed in Section~\ref{subsec:evaluation_methods} the production of
English queries has to be carefully conducted, and is thus expensive.
Besides this, in this paper we do not evaluate our system in terms of
presentation methodology, because experiments using human subjects is
also expensive and still problematic. These remaining issues need to
be further explored.

\section{System Overview}
\label{sec:system_overview}

\begin{figure}[htbp]
  \begin{center}
    \leavevmode
    \psfig{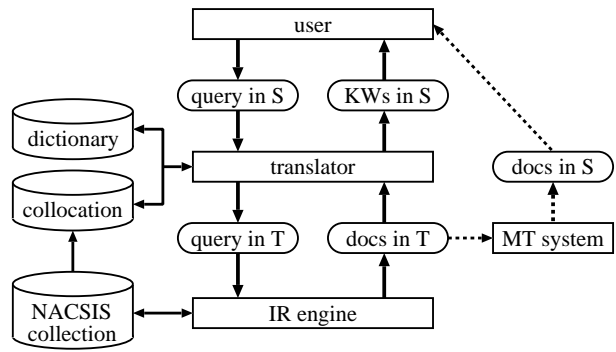}
  \end{center}
  \caption{The overall design of our CLIR system (S and T
  denote the source and target languages, respectively)}
  \label{fig:system}
\end{figure}

Figure~\ref{fig:system} depicts the overall design of our CLIR system,
in which we combine a translator with an IR engine for monolingual
retrieval. In the following, we briefly explain the retrieval process
based on this figure.

First, the translator processes a query in the source language
(query in S) to output the translation (query in T). For this
purpose, the translator uses a dictionary to derive possible
translation candidates and a collocation to resolve the
translation ambiguity. Note that a user can utilize more than one
translation candidate, because multiple translations are often
appropriate for a single query. By the collocation, we mean
bi-gram statistics associated with content words extracted from NACSIS
documents. Since our system is bidirectional between Japanese and
English, we tokenize documents with different methods, depending on
their language. For English documents, the tokenization involves
eliminating stopwords and identifying root forms for inflected content
words. For this purpose, we use
WordNet~\cite{fellbaum:wordnet-98}, which contains a stopword list
and correspondences between inflected words and their root form. On
the other hand, we segment Japanese documents into lexical units using
the ChaSen morphological analyzer~\cite{matsumoto:chasen-97},
which has commonly been used for much Japanese NLP research, and
extract content words based on their part-of-speech information.

Second, the IR engine searches the NACSIS collection for documents
(docs in T) relevant to the translated query, and sorts them
according to the degree of relevance, in descending order. Our IR
engine is currently a simple implementation of the vector space model,
in which the similarity between the query and each document (i.e., the
degree of relevance of each document) is computed as the cosine of the
angle between their associated vectors. We used the notion of
TF$\cdot$IDF for term weighting. Among a number of variations of term
weighting methods~\cite{salton:ipm-88,zobel:sigir-forum-98}, we
tentatively implemented two alternative types of TF (term frequency)
and one type of IDF (inverse document frequency), as shown in
Equation~\eq{eq:tf_idf}.
\begin{equation}
  \label{eq:tf_idf}
  \begin{array}{llll}
    TF & = & f_{t,d} & (\mbox{\rm standard formulation}) \\
    \noalign{\vskip 1.2ex}
    TF & = & 1 + \log(f_{t,d}) & (\mbox{\rm logarithmic formulation})
    \\
    \noalign{\vskip 1.2ex}
    IDF & = & \log(\frac{\textstyle N}{\textstyle n_{t}})
  \end{array}
\end{equation}
Here, $f_{t,d}$ denotes the frequency that term $t$ appears in
document $d$, and $n_{t}$ denotes the number of documents containing
term $t$. $N$ is the total number of documents in the collection. The
second TF type diminishes the effect of $f_{d,t}$, and consequently
IDF affects the similarity computation more. We shall call the first
and second TF types ``standard'' and ``logarithmic'' formulations,
respectively. For the indexing process, we first tokenize documents as
explained above (i.e., we use WordNet and ChaSen for English and
Japanese documents, respectively), and then conduct the word-based
indexing. That is, we use each content word as a single indexing term.
Since our focus in this paper is the query translation rather than the
retrieval process, we do not explore other IR techniques, including
query expansion and relevance feedback.

Finally, in the case where retrieved documents are not in the user's
native language, we extract keywords from retrieved documents, and
translate them into the source language using the translator (KWs in
S). Unlike existing presentation methods, where keywords are words
frequently appearing in each
document~\cite{aone:anlp-97,suzuki:signl-98-7,suzuki:nlp-99}, we
tentatively use author keywords. In the NACSIS collection, each
document contains roughly 3-5 single/compound keywords provided by the
author(s) of the document. In addition, since the NACSIS documents are
relatively short abstracts (instead of entire papers), it is not
entirely satisfactory to rely on the word frequency information. Note
that even in the case where retrieved documents are in the user's
native language, presenting author keywords is expected to improve the
retrieval efficiency.

For future enhancement, we optionally use an MT system to translate
entire documents retrieved (or only documents identified as relevant
using author keywords) into the user's native language (docs in S). We
currently use the Transer Japanese/English MT system, which combines a
general dictionary consisting of 230,000 entries, and a computer
terminology dictionary consisting of 100,000
entries.\footnote{Developed by NOVA, Inc.} Note that the translation
of the limited number of retrieved documents is less expensive than
that of the whole collection, as performed in the document translation
approach (see Section~\ref{subsec:retrieval_methods}).

In Section~\ref{sec:translation}, we will explain the translator
in Figure~\ref{fig:system}, which involves compound word translation
and transliteration methods. While our translation method is
applicable to both queries and keywords in documents, in the following
we shall call it the query translation method without loss of
generality.

\section{Query Translation Method}
\label{sec:translation}

\subsection{Overview}
\label{subsec:trans_overview}

Given a query in the source language, tokenization is first performed
as for target documents, that is, we use WordNet and ChaSen for
English and Japanese queries, respectively (see
Section~\ref{sec:system_overview}). We then discard stopwords and
extract only content words. Here, ``content words'' refer to both
single and compound words. Let us take the following English query as
an example:
\begin{list}{}{}
\item improvement or proposal of data mining methods.
\end{list}
For this query, we discard ``or'' and ``of,'' to extract
``improvement,'' ``proposal'' and ``data mining methods.''
Thereafter, we translate each extracted content word on a word-by-word
basis, maintaining the word order in the source language. A
preliminary study showed that approximately 95\% of compound technical
terms defined in a bilingual dictionary~\cite{ferber:89} maintain the
same word order in both Japanese and English. Note that we currently
do not consider relation (e.g., syntactic relation) between content
words, and thus each content word is translated independently.  In
brief, our translation method consists of the following two phases:
\begin{enumerate}
  \def\labelenumi{(\theenumi)}
\item derive all possible translations for base words,
\item resolve translation ambiguity using the collocation associated
  with base word translations.
\end{enumerate}
While phase~(2) is the same for both Japanese-English and
English-Japanese translations, phase~(1) differs depending on the
source language. In the case of English-Japanese translation, we
simply consult our bilingual dictionary for each base word. However,
transliteration is performed whenever base words unlisted in the
dictionary are found.

On the other hand, in the case of Japanese-English translation, we
consider all possible segmentations of the input word, by consulting
the dictionary, because Japanese compound words lack lexical
segmentation.\footnote{For Japanese query terms used in our evaluation
(see Section~\ref{sec:evaluation}), the average number of possible
segmentations was 4.9.} Then, we select such segmentations that
consist of the minimal number of base words. This segmentation method
parallels that for the Japanese compound noun
analysis~\cite{kobayashi:coling-94}. During the segmentation process,
the dictionary derives all possible translations for base words. At
the same time, transliteration is performed only when {\it katakana\/}
words unlisted in the base word dictionary are found.

\subsection{Compound Word Translation}
\label{subsec:cwt}

This section explains our compound word translation method based on a
probabilistic model, focusing mainly on the resolution of translation
ambiguity. After deriving possible translations for base words (by way
of either consulting the base word dictionary or performing
transliteration), we can formally represent the source compound word
$S$ and one translation candidate $T$ as below.
\begin{eqnarray*}
    S & = & s_{1}, s_{2}, \ldots, s_{n} \\
    T & = & t_{1}, t_{2}, \ldots, t_{n}
\end{eqnarray*}
Here, $s_{i}$ denotes an $i$-th base word, and $t_{i}$ denotes a
translation candidate of $s_{i}$. Our task, i.e., to select the $T$
which maximizes $P(T|S)$, is transformed into
Equation~\eq{eq:trans_model} through use of the Bayesian theorem, as
performed in the statistical machine translation~\cite{brown:cl-93}.
\begin{eqnarray}
  \label{eq:trans_model}
  \arg\max_{T}P(T|S) & = & \arg\max_{T}P(S|T)\cdot P(T)
\end{eqnarray}
In practice, in the case where the user utilizes more than one
translation, $T$'s with greater probabilities are selected. We
approximate $P(S|T)$ and $P(T)$ using statistics associated with base
words, as in Equation~\eq{eq:approx}.
\begin{equation}
  \label{eq:approx}
  \begin{array}{lll}
    P(S|T) & \approx & {\displaystyle \prod_{i=1}^{n}P(s_{i}|t_{i})} \\
    \noalign{\vskip 1.2ex}
    P(T) & \approx & {\displaystyle
    \prod_{i=1}^{n-1}P(t_{i+1}|t_{i})}
  \end{array}
\end{equation}
One may notice that this approximation is analogous to that for the
statistical part-of-speech tagging, where $s_{i}$ and $t_{i}$ in
Equation~\eq{eq:approx} correspond to a word and one of its
part-of-speech candidates, respectively~\cite{church:cl-93}. Here, we
estimate $P(t_{i+1}|t_{i})$ using the word-based bi-gram statistics
extracted from target language documents (i.e., the collocation in
Figure~\ref{fig:system}). Before elaborating on the estimation of
$P(s_{i}|t_{i})$ we explain the way to produce our bilingual
dictionary for base words, because $P(s_{i}|t_{i})$ is estimated using
this dictionary.

For our dictionary production, we used the EDR technical terminology
dictionary~\cite{edr-techdic:95}, which includes approximately 120,000
Japanese-English translations related to the information processing
field. Since most of the entries are compound words, we need to
segment Japanese compound words, and correlate Japanese-English
translations on a word-by-word basis.  However, the complexity of
segmenting Japanese words becomes much greater as the number of
component base words increases. In consideration of these factors, we
first extracted 59,533 English words consisting of only {\em two\/}
base words, and their Japanese translations. We then developed simple
heuristics to segment Japanese compound words into two substrings. Our
heuristics relies mainly on Japanese character types, i.e., ``{\it
kanji},'' ``{\it katakana},'' ``{\it hiragana},'' alphabets and other
characters like numerals. Note that {\it kanji\/} (or Chinese
character) is the Japanese idiogram, and {\it katakana\/} and {\it
hiragana\/} are phonograms.

In brief, we segment each Japanese word at the boundary of different
character types (or at the leftmost boundary for words containing more
than one character type boundary). Although this method is relatively
simple, a preliminary study showed that we can almost correctly
segment words that are in one of the following forms: ``{\tt CK},''
``{\tt CA},'' ``{\tt AK}'' and ``{\tt KA\/}.'' Here, ``{\tt C},''
``{\tt K\/}'' and ``{\tt A}'' denote {\it kanji}, {\it katakana\/} and
alphabet character sequences, respectively. For other combinations of
character types, we identified one or more cases in which our
segmentation method incorrectly performed.

On the other hand, in the case where a given Japanese word consists of
a single character type, we segment the word at the middle (or at the
left-side of the middle character for words consisting of an odd
number of characters).  Note that roughly 90\% of Japanese words
consisting of four {\it kanji\/} characters can be correctly segmented
at the middle~\cite{kobayashi:coling-94}. However, in the case where
resultant substrings begin/end with characters that do not appear at
the beginning/end of words (for example, Japanese words rarely begin
with a long vowel), we shift the segmentation position to the right.

Tsuji and Kageura~\nocite{tsuji:nlprs-97} used the HMM to segment
Japanese compound words in an English-Japanese bilingual
dictionary. Their method can also segment words consisting of more
than two base words, and reportedly achieved an accuracy of roughly
80-90\%, whereas our segmentation method is applicable only to those
consisting of two base words. However, while the HMM-based
segmentation is expected to improve the quality of our dictionary
production, in this paper we tentatively show that our
heuristics-based method is effective for CLIR despite its simple
implementation, by way of experiments (see
Section~\ref{sec:evaluation}).

As a result, we obtained 24,439 Japanese and 7,910 English base words.
We randomly sampled 600 compound words, and confirmed that 95\% of
those words were correctly segmented.
Figure~\ref{fig:compound_word_dictionary} shows a fragment of the EDR
dictionary (after segmenting Japanese words), and
Figure~\ref{fig:base_word_dictionary} shows a base word dictionary
produced from entries in Figure~\ref{fig:compound_word_dictionary}.
Figure~\ref{fig:base_word_dictionary} contains Japanese variants, such
as {\it memori\/}/{\it memorii\/} for the English word ``memory.'' We
can easily produce a Japanese-English base word dictionary from
Figure~\ref{fig:compound_word_dictionary}, using the same procedure.

During the dictionary production, we also count the correspondence
frequency for each combination of $s_{i}$ and $t_{i}$, in order to
estimate $P(s_{i}|t_{i})$. In Figure~\ref{fig:base_word_dictionary},
for example, the Japanese base word ``{\it soukan\/}'' corresponds
once to ``associative,'' and twice to ``correlation.'' Thus, we can
derive Equation~\eq{eq:soukan}.
\begin{equation}
  \label{eq:soukan}
  \begin{array}{lll}
    P(\mbox{associative}\:|\:{\it soukan}) & = & 1/3 \\ 
    \noalign{\vskip 0.6ex}
    P(\mbox{correlation}\:|\:{\it soukan}) & = & 2/3
  \end{array}
\end{equation}
However, in the case where $s_{i}$ is {\em transliterated\/} into
$t_{i}$, we replace $P(s_{i}|t_{i})$ with a probabilistic score
computed by our transliteration method (see
Section~\ref{subsec:translit}).

One may argue that $P(s_{i}|t_{i})$ should be estimated based on real
world usage, i.e., bilingual corpora. However, such resources are
generally expensive to obtain, and we do not have Japanese-English
corpora with sufficient volume of alignment information at present
(see Section~\ref{subsec:our_approach} for more discussion).

\begin{figure}[htbp]
  \def\baselinestretch{1}
  \begin{center}
    \leavevmode
    \small
    \begin{tabular}[t]{ll} \hline\hline
      {\hfill\centering English\hfill} & {\hfill\centering
      Japanese\hfill} \\ \hline
      CCD memory & CCD {\it memorii\/} \\
      IC memory & IC {\it memori\/} \\
      associative learning & {\it soukan gakushuu\/} \\
      associative memory & {\it rensou memori\/} \\
      associative record & {\it ketsugou rekoodo\/} \\
      correlation function & {\it soukan kansuu\/} \\
      error detection & {\it ayamari kenshutsu\/} \\
      factor correlation & {\it inshi soukan\/} \\
      hybrid IC & {\it haiburiddo shuusekikairo\/} \\ \hline
    \end{tabular}
  \end{center}
  \caption{A fragment of the EDR technical terminology dictionary}
  \label{fig:compound_word_dictionary}
\end{figure}

\subsection{Transliteration}
\label{subsec:translit}

This section explains our transliteration method, which identifies
phonetic equivalent translations for words unlisted in the base word
dictionary.

Figure~\ref{fig:katakana} shows example correspondences between
English and (romanized) {\it katakana\/} words, where we insert
hyphens between each {\it katakana\/} character for enhanced
readability. The basis of our transliteration method is analogous to
that for compound word translation described in
Section~\ref{subsec:cwt}. The formula for the source word $S$ and one
transliteration candidate $T$ are represented as below.
\begin{eqnarray*}
    S & = & s_{1}, s_{2}, \ldots, s_{n} \\
    T & = & t_{1}, t_{2}, \ldots, t_{n}
\end{eqnarray*}
Here, unlike the case of compound word translation, $s_{i}$ and
$t_{i}$ denote $i$-th ``symbols'' (which consist of one or more
letters), respectively. To derive possible $s_{i}$'s and $t_{i}$'s, we
consider all possible segmentations of the source word $S$, by
consulting a dictionary for symbols, namely the ``transliteration
dictionary.'' Then, we select such segmentations that consist of the
minimal number of symbols. Note that unlike the case of compound word
translation, the segmentation is performed for both Japanese-English
and English-Japanese transliterations.

\begin{figure}[htbp]
  \def\baselinestretch{1}
  \begin{center}
    \leavevmode
    \small
    \begin{tabular}[t]{ll} \hline\hline
      {\hfill\centering English\hfill} & {\hfill\centering
      Japanese\hfill} \\ \hline
      CCD & CCD \\
      IC & IC, {\it shuusekikairo\/} \\
      associative & {\it soukan}, {\it rensou}, {\it ketsugou\/} \\
      correlation & {\it soukan\/} \\
      detection & {\it kenshutsu\/} \\
      error & {\it ayamari\/} \\
      factor & {\it inshi\/} \\
      function & {\it kansuu\/} \\
      hybrid & {\it haiburiddo\/} \\
      learning & {\it gakushuu\/} \\
      memory & {\it memori}, {\it memorii\/} \\
      record & {\it rekoodo\/} \\ \hline
    \end{tabular}
  \end{center}
  \caption{A fragment of an English-Japanese base word dictionary
  produced from Figure~\protect\ref{fig:compound_word_dictionary}}
  \label{fig:base_word_dictionary}
\end{figure}

\begin{figure}[htbp]
  \def\baselinestretch{1}
  \begin{center}
    \leavevmode
    \small
    \begin{tabular}{ll} \hline\hline
      {\hfill\centering English \hfill} & {\hfill\centering Japanese
      \hfill} \\ \hline
      system & {\it shi-su-te-mu\/} \\
      mining & {\it ma-i-ni-n-gu\/} \\
      data & {\it dee-ta\/} \\
      network & {\it ne-tto-waa-ku\/} \\
      text & {\it te-ki-su-to\/} \\
      collocation & {\it ko-ro-ke-i-sho-n\/} \\ \hline
    \end{tabular}
    \caption{Example correspondences between English and  (romanized)
    Japanese {\it katakana\/} words}
    \label{fig:katakana}
  \end{center}
\end{figure}

Thereafter, we resolve the transliteration ambiguity based on the a
probabilistic model similar to that for the compound word translation.
To put it more precisely, we compute $P(T|S)$ for each $T$ using
Equation~\eq{eq:trans_model}, and select $T$'s with greater
probabilities. Note that $T$'s must be correct words (that are indexed
in the NACSIS document collection).  However, Equation~\eq{eq:approx},
which approximates $P(T)$ by combining $P(t_i)$'s for substrings of
$T$, potentially assigns positive possibility values for incorrect
(unindexed) words.

In view of this problem, we estimate $P(T)$ as the probability that
$T$ occurs in the document collection, and consequently the
probability for unindexed words becomes zero. In practice, during the
segmentation process we simply discard such $T$'s that are unindexed
in the document collection, so that we can enhance the computation for
$P(T|S)$'s.  On the other hand, we approximate $P(S|T)$ as in
Equation~\eq{eq:approx}, and estimate $P(s_{i}|t_{i})$ based on the
correspondence frequency for each combination of $s_{i}$ and $t_{i}$
in the transliteration dictionary.

The crucial content here is the way to produce the transliteration
dictionary, because such dictionaries have rarely been published. For
the purpose of dictionary production, we used approximately 35,000
{\it katakana\/} Japanese words and their English translations
collected from the EDR technical terminology
dictionary~\cite{edr-techdic:95} and bilingual
dictionary~\cite{edr-bilindic:95}. To illustrate our dictionary
production method, we consider Figure~\ref{fig:katakana}
again. Looking at this figure, one may notice that the first letter in
each {\it katakana\/} character tends to be contained in its
corresponding English word. However, there are a few exceptions. A
typical case is that since Japanese has no distinction between ``L''
and ``R'' sounds, the two English sounds collapse into the same
Japanese sound. In addition, a single English letter may correspond to
multiple {\it katakana\/} characters, such as ``x'' to ``{\it
ki-su\/}'' in \mbox{``$<$text, {\it te-ki-su-to\/}$>$.''} To sum up,
English and romanized {\it katakana\/} words are not exactly
identical, but similar to each other.

\begin{figure}[htbp]
  \begin{center}
    \leavevmode
    \psfig{file=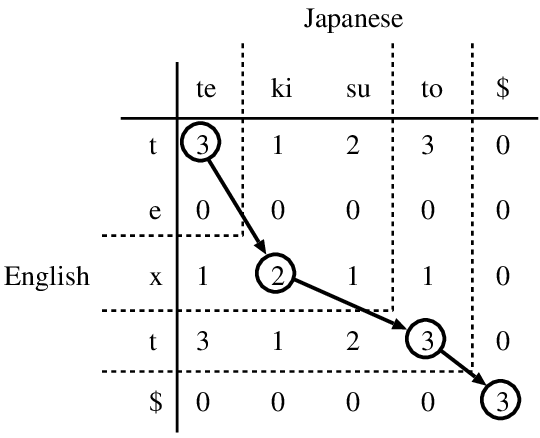,height=2in}
  \end{center}
  \caption{An example matrix for English-Japanese symbol matching
  (arrows denote the best path)}
  \label{fig:matrix}
\end{figure}

We first manually defined the similarity between the English letter
$e$ and the first romanized letter for each {\it katakana\/} character
$j$, as shown in Table~\ref{tab:katakana}. In this table,
``phonetically similar'' letters refer to a certain pair of letters,
such as ``L'' and ``R,'' for which we identified approximately twenty
pairs of letters. We then consider the similarity for any possible
combination of letters in English and romanized {\it katakana\/}
words, which can be represented as a matrix, as shown in
Figure~\ref{fig:matrix}. This figure shows the similarity between
letters in \mbox{``$<$text, {\it te-ki-su-to\/}$>$.''}  We put a dummy
letter ``\$,'' which has a positive similarity only to itself, at the
end of both English and {\it katakana\/} words.

One may notice that matching plausible symbols can be seen as finding
the path which maximizes the total similarity from the first to last
letters. The best path can efficiently be found by, for example,
Dijkstra's algorithm~\cite{dijkstra:nm-59}. From
Figure~\ref{fig:matrix}, we can derive the following correspondences:
\mbox{``$<$te, {\it te\/}$>$,''} \mbox{``$<$x, {\it ki-su\/}$>$''} and
\mbox{``$<$t, {\it to\/}$>$.''} In practice, to exclude noisy
correspondences, we used only English-Japanese translations whose
total similarity from the first to last letters is above a predefined
threshold. The resultant transliteration dictionary contains 432
Japanese and 1018 English symbols, from which we estimated
$P(s_{i}|t_{i})$.

\begin{table}[htbp]
  \def\baselinestretch{1}
  \begin{center}
    \caption{The similarity between English letter $e$ and Japanese
    letter $j$}
    \medskip \leavevmode \small
    \begin{tabular}{lc} \hline\hline
      {\hfill\centering Condition \hfill} & {\hfill\centering
      Similarity \hfill} \\ \hline
      $e$ and $j$ are identical & 3 \\
      $e$ and $j$ are phonetically similar & 2 \\
      both $e$ and $j$ are vowels or consonants & 1 \\
      otherwise & 0 \\ \hline
    \end{tabular}
    \label{tab:katakana}
  \end{center}
\end{table}

To evaluate our transliteration method, we extracted Japanese {\it
katakana\/} words (excluding compound words) and their English
translations from an English-Japanese
dictionary~\cite{nichigai_compdic:96}. We then discarded
Japanese/English pairs that were not phonetically equivalent to each
other, and were listed in the EDR dictionaries. For the resultant 248
pairs, the accuracy of our transliteration method was 65.3\%.

Thus, our transliteration method is less accurate than the word-based
translation. For example, the {\it katakana\/} word ``{\it
re-ji-su-ta}~(register/resistor)'' is transliterated into
``resister,'' ``resistor'' and ``register,'' with the probability
score in descending order. Note that Japanese seldom represents
``resister'' as ``{\it re-ji-su-ta\/}'' (whereas it can be
theoretically correct when this word is written in {\it katakana\/}
characters), because ``resister'' corresponds to more appropriate
translations in {\it kanji\/} characters. However, the compound word
translation is expected to select appropriate transliteration
candidates. For example, ``re-ji-su-ta'' in the compound word ``{\it
re-ji-su-ta\/} {\it tensou\/} {\it gengo\/}~(register transfer
language)'' is successfully translated, given a set of base words
``{\it tensou\/}~(transfer)'' and ``{\it gengo\/}~(language)'' as a
context.

Finally, we devote a little more space to compare our transliteration
method and other related works.
Chen~\etal~\nocite{chen:coling-acl-98} proposed a Chinese-English
transliteration method. Given a (romanized) source word, their methods
compute the similarity between the source word and each target word
listed in the dictionary. In brief, the more letters two words share
in common, the more similar they are. In other words, unlike our case,
their methods disregard the order of letters in source and target
words, which potentially degrades the transliteration accuracy. In
addition, since for each source word the similarity is computed
between all the target words (or words that share at least one common
letter with the source word), the similarity computation can be
prohibitive. Lee and Choi~\nocite{lee:iral-97} explored English-Korean
transliteration, where they automatically produced a transliteration
model from a word-aligned corpus. In brief, they first consider all
possible English-Korean symbol correspondences for each word
alignment. Then, iterative estimation is performed to select such
symbol correspondences that maximize transliteration accuracy on
training data. However, when compared with our symbol alignment
method, their iterative estimation method is computationally
expensive. Knight and Graehl~\nocite{knight:cl-98} proposed a
Japanese-English transliteration method based on the mapping
probability between English and Japanese {\it katakana\/}
sounds. However, while their method needs a large-scale phoneme
inventory, we use a simpler approach using surface mapping between
English and {\it katakana\/} characters, as defined in our
transliteration dictionary. Note that none of those above methods has
been evaluated in the context of CLIR. Empirical comparison of
different transliteration methods needs to be further explored.

\subsection{Further Enhancement of Translation}
\label{subsec:dictionary_enhancement}

This section explains two additional methods to enhance the query
translation.

First, we can enhance our base word dictionary with {\em general\/}
words, because technical compound words sometimes include general
words, as discussed in Section~\ref{sec:introduction}. Note that in
Section~\ref{subsec:cwt} we produced our base word dictionary from the
EDR {\em technical\/} terminology dictionary. Thus, we used the EDR
bilingual dictionary~\cite{edr-bilindic:95}, which consists of
approximately 370,000 Japanese-English translations aimed at general
usage. However, unlike in the case of technical terms, it is not
feasible to segment general compound words, such as ``hot dog,'' into
base words. Thus, we simply extracted 162,751 Japanese and 67,136
English single words (i.e., words that consist of a single base word)
from this dictionary.  In addition, to minimize the degree of
translation ambiguity, we use general translations only when (a) base
words unlisted in our technical term dictionary are found, and (b) our
transliteration method fails to output any candidates for those
unlisted base words.

Second, in Section~\ref{sec:introduction} we also identified that
English technical terms are often abbreviated, such as ``IR'' and
``NLP,'' and they can be used as Japanese words. One solution would be
to output those abbreviated words as they are, for both
Japanese-English and English-Japanese translations. On the other hand,
it is expected that we can improve the recall by using complete forms
along with their abbreviated forms. To realize this notion, we
extracted 7,307 tuples of each abbreviation and its complete form from
the NACSIS English document collection, using simple heuristics. Our
heuristics relies on the assumption that either abbreviations or
complete forms often appear in parentheses headed by their
counterparts, as shown below:
\begin{quote}
  Natural Language Processing (NLP), \\
  cross-language information retrieval (CLIR), \\
  MRDs (machine readable dictionaries).
\end{quote}
While the first example is the most straightforward, in the second and
third examples we disregard a hyphen and lowercase letter (i.e., ``s''
in ``MRDs''), respectively. In practice, we can easily extract such
tuples using the regular expression pattern matching.
Figure~\ref{fig:abbreviation} shows example tuples of abbreviations
and complete forms extracted from the NACSIS collection.  In this
figure, the column ``Frequency'' denotes the frequency that each tuple
appears in the collection, with which we can optionally set a cut-off
threshold for multiple complete forms corresponding to a single
abbreviation (e.g., ``information retrieval,'' ``isoprene rubber'' and
``insulin receptor'' for ``IR'').

\begin{figure}[htbp]
  \def\baselinestretch{1}
  \begin{center}
    \leavevmode
    \small
    \begin{tabular}[t]{llc} \hline\hline
      {\hfill\centering Abbreviation\hfill} & {\hfill\centering
      Complete form\hfill} & {\hfill\centering Frequency\hfill} \\ \hline
      IR & information retrieval & 3 \\
      IR & isoprene rubber & 1 \\
      IR & insulin receptor & 1 \\
      MT & machine translation & 11 \\
      MT & mobile telephone & 3 \\
      NLP & natural language processing & 8 \\ \hline
    \end{tabular}
  \end{center}
  \caption{Example abbreviations and their complete forms}
  \label{fig:abbreviation}
\end{figure}

\section{Evaluation}
\label{sec:evaluation}

\subsection{Methodology}
\label{subsec:eval_overview}

We investigated the performance of our system in terms of
Japanese-English CLIR, based on the TREC-type evaluation methodology.
That is, the system outputs 1,000 top documents, and the TREC
evaluation software was used to plot recall-precision curves and
calculate non-interpolated average precision values.

For the purpose of our evaluation, we used a preliminary version of
the NACSIS test collection~\cite{kando:sigir-99}. This collection
includes approximately 330,000 documents (in either a combination of
English and Japanese or either of the languages individually),
collected from technical papers published by 65 Japanese associations
for various fields.\footnote{The official version of the NACSIS
collection includes 39 Japanese queries and the same document set as
in the preliminary version we used. NACSIS (National Center for
Science Information Systems, Japan) held a TREC-type (CL)IR contest
workshop in August 1999, and participants, including the authors of
this paper, were provided with the whole document set and 21 queries
for training. These 21 queries are included in the final package of
the test collection. See {\tt
http://www.rd.nacsis.ac.jp/\~{}ntcadm/workshop/work-en.html} for
details.} Each document consists of the document ID, title, name(s) of
author(s), name/date of conference, hosting organization, abstract and
keywords, from which we used titles, abstracts and keywords for the
indexing. We used as target documents approximately 187,000 entries
where abstracts are in both English and Japanese.

This collection also includes 21 Japanese queries. Each query
consists of the query ID, title of the topic, description, narrative
and list of synonyms, from which we used only the
description.\footnote{In the NACSIS workshop, each participant can
submit more than one retrieval result using different
systems. However, at least one result must be gained with only the
description field.} In general, most topics are related to electronic,
information and control engineering. Figure~\ref{fig:query} shows
example descriptions (translated into English by one of the authors).

In the NACSIS collection, relevance assessment was performed based on
the pooling method~\cite{voorhees:sigir-98}. That is, candidates for
relevant documents were first obtained with multiple retrieval
systems. Thereafter, for each candidate document, human expert(s)
assigned one of three ranks of relevance, i.e., ``relevant,''
``partially relevant'' and \mbox{``irrelevant.''} The average number
of candidate documents for each query is 4,400, among which the number
of relevant and partially relevant documents are 144 and 13,
respectively. In our evaluation, we did not regard partially relevant
documents as relevant ones, because (a) the result did not
significantly change depending on whether we regarded partially
relevant as relevant or not, and (b) interpretation of partially
relevant is not fully clear to the authors.

Since the NACSIS collection does not contain English queries, we
cannot estimate a baseline for Japanese-English CLIR performance based
on English-English IR. Instead, we used a Japanese-Japanese IR system,
which uses as documents Japanese titles/abstracts/keywords comparable
to English fields in the NACSIS collection.  One may argue that we can
manually translate Japanese queries into English. However, as
discussed in Section~\ref{subsec:evaluation_methods}, the CLIR
performance varies depending on the quality of translation, and thus
we avoided an arbitrary evaluation.

\begin{figure}[htbp]
  \def\baselinestretch{1}
  \begin{center}
    \leavevmode
    \small
    \begin{tabular}{cl} \hline\hline
      ID & {\hfill\centering Description\hfill} \\ \hline
      0005 & dimension reduction for clustering \\
      0006 & intelligent information retrieval using agent functions \\
      0019 & syntactic analysis methods for Japanese sentences \\
      0024 & machine translation systems \\ \hline
    \end{tabular}
    \caption{Example query descriptions in the NACSIS collection}
    \label{fig:query}
  \end{center}
\end{figure}

\subsection{Quantitative Comparison}
\label{subsec:quantitative}

We compared the following query translation methods:
\begin{itemize}
\item all possible translations derived from the (original) EDR
  technical terminology dictionary~\cite{edr-techdic:95} are used for
  query terms, which can be seen as a lower bound method of this
  comparative experiment (``EDR''),
\item all possible base word translations derived from our base word
  dictionary are used (``ALL''),
\item $k$-best translations selected by our compound word translation
  method are used, where transliteration is not used (``CWT''),
\item transliteration is performed for unlisted {\it katakana\/} words
  in CWT above, which represents the overall query
  translation method we proposed in this paper (``TRL'').
\end{itemize}
One may notice that both EDR and ALL correspond to the
dictionary-based method, and CWT and TRL correspond to the
hybrid method described in Section~\ref{subsec:retrieval_methods}. In
the case of EDR, compound words unlisted in the EDR dictionary
were manually segmented so that substrings (shorter compound words or
base words) could be translated. There was almost no translation
ambiguity in the case of EDR. In addition, preliminary experiments
showed that disambiguation degraded the retrieval performance for
EDR. In CWT and TRL, $k$ is a parametric constant, for
which we set \mbox{$k=1$}. Through preliminary experiments, we
achieved the best performance when we set \mbox{$k=1$}. By increasing
the value of $k$, we theoretically gain a query expansion effect,
because multiple translations semantically related are used as query
terms. However, in our case, additional translations were rather noisy
with respect to the retrieval performance. Note that in this
experiment, we did not used the general and abbreviation dictionaries.
We will discuss the effect of those dictionaries in
Section~\ref{subsec:dictionary_enhancement}.

Table~\ref{tab:avg_pre} shows the non-interpolated average precision
values, averaged over the 21 queries, for different combinations of
query translation and retrieval methods. It is worth comparing the
effectiveness of query translation methods with different retrieval
methods, because advanced retrieval methods potentially overcome the
rudimentary nature of query translation methods, and therefore may
overshadow the difference of query translation methods in CLIR
performance. In consideration of this problem, as described in
Section~\ref{sec:system_overview}, we adopted two alternative term
weighting methods, i.e., the standard and logarithmic formulations. In
addition, we used as the IR engine in Figure~\ref{fig:system} the
SMART system~\cite{salton:71}, where the augmented TF$\cdot$IDF term
weighting method (``ATC'') was used for both queries and
documents. This makes it easy for other researchers to rigorously
compare their query translation methods with ours within the same
evaluation environment, because the SMART system is available to the
public.

In Table~\ref{tab:avg_pre}, J-J refers to the baseline performance,
that is, the result obtained by the Japanese-Japanese IR system.  Note
that the performance of J-J using the SMART system is not available
because this system is not implemented for the retrieval of Japanese
documents. The column ``\# of Terms'' denotes the average number of
query terms used for the retrieval, where the number of terms used in
ALL was approximately seven times as great as those of other
methods. Suggestions can be derived from these results is as follows.

\begin{table}[htbp]
  \def\baselinestretch{1}
  \begin{center}
    \caption{Non-interpolated average precision values,
    averaged over the 21 queries, for different combinations of query
    translation and retrieval methods}
    \medskip
    \leavevmode
    \small
    \begin{tabular}{lccccc} \hline\hline
      & & \multicolumn{3}{c}{Retrieval Method} \\ \cline{3-5}
      & \# of Terms & Standard TF & Logarithmic TF & SMART \\ \hline
      J-J & 4.0 & 0.2085 & 0.2443 & --- \\
      TRL & 4.0 & 0.2427 & 0.2911 & 0.3147 \\
      CWT & 3.9 & 0.2324 & 0.2680 & 0.2770 \\
      ALL & 21  & 0.1971 & 0.2271 & 0.2106 \\
      EDR & 4.1 & 0.1785 & 0.2173 & 0.2477 \\ \hline
    \end{tabular}
    \label{tab:avg_pre}
  \end{center}
\end{table}

First, the relative superiority between EDR and ALL varies
depending on the retrieval method. Since neither case resolved the
translation ambiguity, the difference in performance for the two
translation methods is reduced solely to the difference between the
two dictionaries. Therefore, the base word dictionary we produced was
effective when combined with the standard and logarithmic TF
formulations. However, the translation disambiguation as performed in
CWT improved the performance of ALL, and consequently CWT
outperformed EDR irrespective of the retrieval method. To sum up,
our compound word translation method was more effective than the use
of an existing dictionary, in terms of CLIR performance.

Second, by comparing results of CWT and TRL, one can see that
our transliteration method further improved the performance of the
compound word translation relying solely on the base word dictionary,
irrespective of the retrieval method.  Since TRL represents the
overall performance of our system, it is worth comparing TRL and
EDR (i.e., a lower bound method) more carefully. Thus, we used the
paired t-test for statistical testing, which investigates whether the
difference in performance is meaningful or simply due to
chance~\cite{hull:sigir-93,keen:ipm-92}. We found that the average
precision values of TRL and EDR are significantly different
(at the 5\% level), for any of the three retrieval methods.

Third, the performance was generally improved as a more sophisticated
retrieval method was used, for all of the translation methods
excepting ALL. In other words, enhancements of the query
translation and IR engine independently improved on the performance of
our CLIR system. Note that the difference between the SMART system and
the other two methods is due to more than one factor, including
stemming and term weighting methods. This suggests that our system may
achieve a higher performance using other advanced IR techniques.

Finally, TRL and CWT outperformed J-J for any of the
retrieval methods. However, these differences are partially attributed
to the different properties inherent in Japanese and English IR. For
example, the performance of Japanese IR is more strongly dependent on
the indexing method than English IR, since Japanese lacks lexical
segmentation. This issue needs to be further explored.

\begin{figure}[htbp]
  \begin{center}
    \leavevmode
    \psfig{file=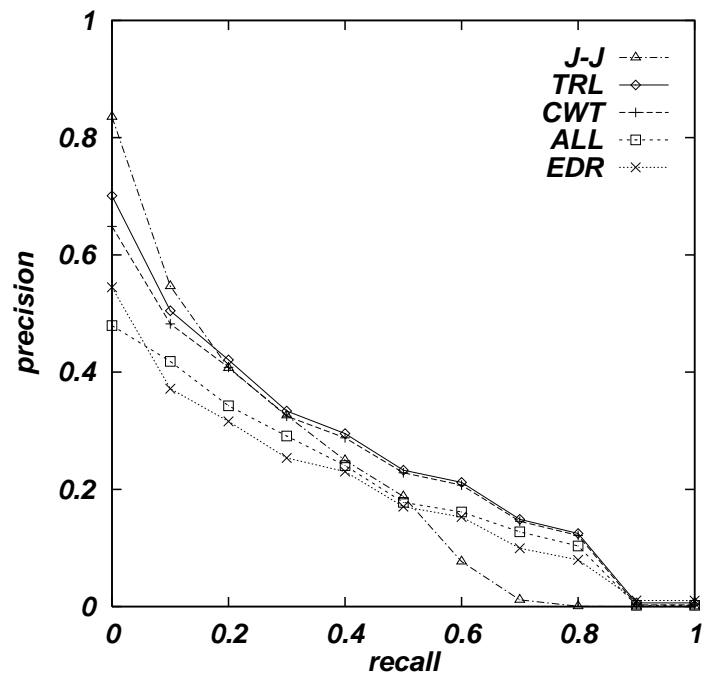,height=3.5in}
  \end{center}
  \caption{Recall-precision curves using the standard TF}
  \label{fig:rp_raw_TF}
\end{figure}

\begin{figure}[htbp]
  \begin{center}
    \leavevmode
    \psfig{file=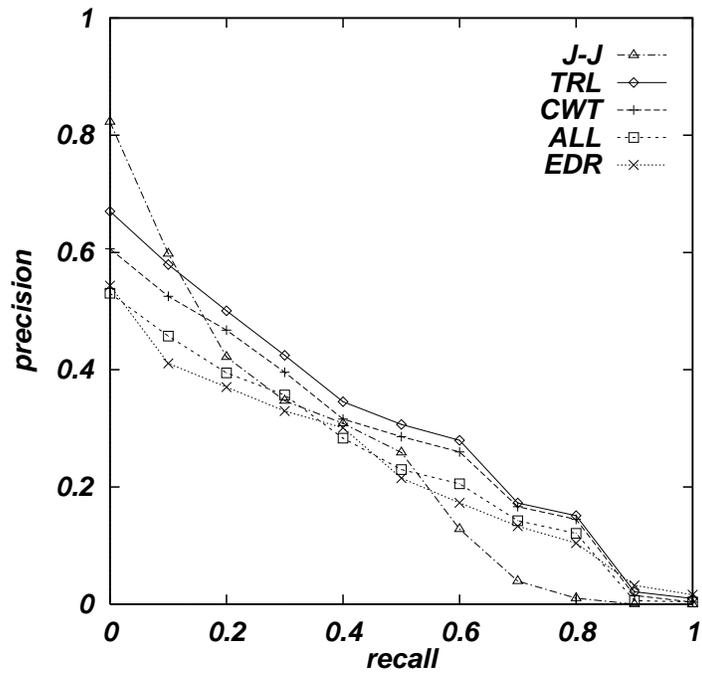,height=3.5in}
  \end{center}
  \caption{Recall-precision curves using the logarithmic TF}
  \label{fig:rp_log_TF}
\end{figure}

\begin{figure}[htbp]
  \begin{center}
    \leavevmode
    \psfig{file=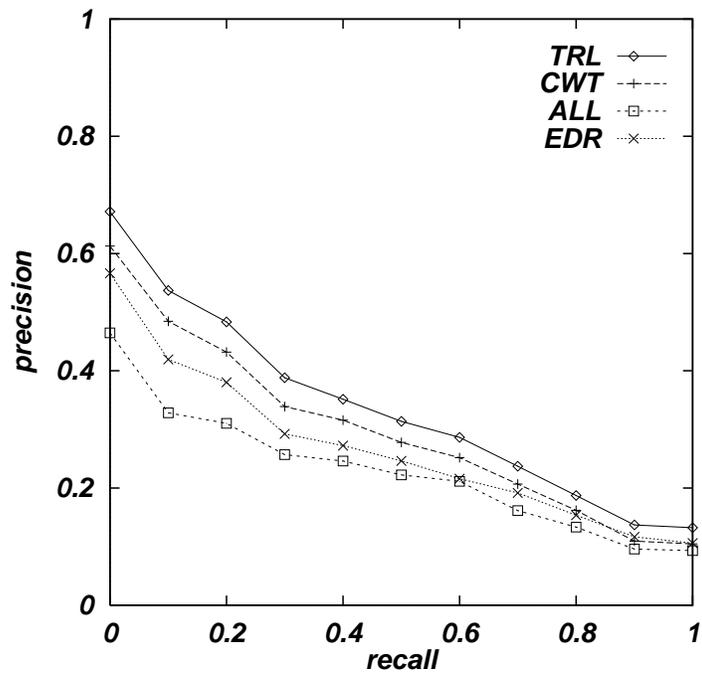,height=3.5in}
  \end{center}
  \caption{Recall-precision curves using the SMART system}
  \label{fig:rp_smart}
\end{figure}

Figures~\ref{fig:rp_raw_TF}-\ref{fig:rp_smart} show recall-precision
curves of different query translation methods, for different retrieval
methods, respectively. In these figures, while the superiority of EDR
and ALL in terms of precision varies depending on the recall, one can
see that CWT outperformed EDR and ALL, and that TRL outperformed CWT,
regardless of the recall. In Figures~\ref{fig:rp_raw_TF} and
\ref{fig:rp_log_TF}, J-J generally performed better at lower recall
while any of four CLIR methods performs better at higher recall. As
discussed above, possible rationales would include the difference
between Japanese and English IR. To put it more precisely, in Japanese
IR a word-based indexing method (as performed in our IR engine) fails
to retrieve documents in which words are inappropriately segmented.
In addition, the ChaSen morphological analyzer often incorrectly
segments {\it katakana\/} words, which frequently appear in technical
documents. Consequently this drawback leads to a poor recall in the
case of J-J.

\subsection{Query-by-query Analysis}
\label{subsec:qbq_analysis}

In this Section, we discuss reasons why our translation method
was effective in CLIR performance, through a query-by-query analysis.

First, we compared EDR and CWT (see in
Section~\ref{subsec:quantitative}), to investigate the effectiveness
of our compound word translation method. For this purpose, we
identified fragments of the NACSIS query that were correctly
translated by CWT but not by EDR, as shown in
Table~\ref{tab:avgpre_qbq_cwt}. In this table, where we insert hyphens
between each Japanese base word for enhanced readability,
Japanese/English words unlisted in the EDR technical terminology
dictionary are underlined. Note that as mentioned in
Section~\ref{subsec:quantitative}, in these cases translations for
remaining base words were used as query terms. However, in the case of
the query 0019, the EDR dictionary lists a phrase translation,
i.e., ``{\it kakariuke-kaiseki\/}~(analysis of dependence relation),''
and thus ``analysis,'' ``dependence'' and ``relation'' were used as
query terms (``of'' was discarded as a stopword).  One can see that
except for the five cases asterisked, out of 18 cases, CWT
outperformed EDR. Note that in the case of 0019, EDR
conducted a phrase-based translation, while CWT conducted a
word-based translation. The relative superiority between these two
translation approaches varies depending on the retrieval method, and
thus we cannot draw any conclusion regarding this point in this paper.
In the case of the query 0006, although the translation in CWT
was linguistically correct, we found that the English word ``agent
function'' is rarely used in documents associated with agent research,
and that ``function'' ended up degrading the retrieval performance. In
the case of the query 0020, ``loanword'' would be a more
appropriate translation for ``{\it gairaigo\/}.'' However, even when
we used ``loanword'' for the retrieval, instead of ``foreign'' and
``word,'' the performance of EDR did not change.

\begin{table}[htbp]
  \def\baselinestretch{1}
  \begin{center}
    \caption{Query-by-query comparison between EDR and CWT}
    \medskip
    \leavevmode
    \footnotesize
    \tabcolsep=3pt
    \begin{tabular}{cllll} \hline\hline
      & & \multicolumn{3}{c}{Change in Average Precision (EDR
      $\rightarrow$ CWT)} \\ \cline{3-5}
      ID & {\hfill\centering Japanese (Translation in CWT)\hfill} &
      {\hfill\centering Standard TF\hfill} & {\hfill\centering
      Logarithmic TF\hfill} & {\hfill\centering SMART\hfill} \\ \hline
      0001 & {\it $\underline{jiritsu}$-idou-robotto\/}
      ($\underline{\mbox{autonomous}}$ mobile robot) & 0.2325
      $\rightarrow$ 0.3667 & 0.2587 $\rightarrow$ 0.4058 & 0.2259
      $\rightarrow$ 0.3441 \\
      0004 & {\it $\underline{bunsho}$-gazou-rikai\/}
      ($\underline{\mbox{document}}$ image understanding) & 0.0011
      $\rightarrow$ 0.2775 & 0.0091 $\rightarrow$ 0.3768 & 0.0217
      $\rightarrow$ 0.2740 \\
      0006 & {\it eejento-$\underline{kinou}$\/} (agent
      $\underline{\mbox{function}}$) & 0.2008
      $\rightarrow$ 0.1603* & 0.2920 $\rightarrow$ 0.1997* & 0.1430
      $\rightarrow$ 0.1395* \\
      0016 & {\it saidai-$\underline{kyoutsuu}$-bubungurafu\/}
      (greatest $\underline{\mbox{common}}$ subgraph) &
      0.1615 $\rightarrow$ 0.5039 & 0.4661 $\rightarrow$ 0.6216 &
      0.1295 $\rightarrow$ 0.4460 \\
      0019 & {\it kakariuke-kaiseki\/} (dependency analysis) & 0.0794
      $\rightarrow$ 0.3550 & 0.1383 $\rightarrow$ 0.4302 & 0.1852
      $\rightarrow$ 0.1449* \\
      0020 & {\it katakana-$\underline{\mbox{\it gairai-go\/}}$\/}
      (katakana $\underline{\mbox{foreign word}}$) & 0.4536
      $\rightarrow$ 0.4568 & 0.2408 $\rightarrow$ 0.4674 & 0.9429
      $\rightarrow$ 0.8769* \smallskip \\ \hline
    \end{tabular}
    \label{tab:avgpre_qbq_cwt}
  \end{center}
\end{table}

Second, we compared CWT and TRL in Table~\ref{tab:avgpre_qbq_trl},
which uses the same basic notation as Table~\ref{tab:avgpre_qbq_cwt}.
The NACSIS query set contains 20 {\it katakana\/} base word types,
among which ``{\it ma-i-ni-n-gu\/}~(mining)'' and ``{\it
ko-ro-ke-i-sho-n\/}~(collocation)'' were unlisted in our base word
dictionary. Unlike the previous case, transliteration generally
improved on the performance. On the other hand, we concede that only
three queries are not enough to justify the effectiveness of our
transliteration method. In view of this problem, we assumed that every
{\it katakana\/} word in the query is unlisted in our base word
dictionary, and compared the following two extreme cases:
\begin{itemize}
\item every {\it katakana\/} word was untranslated (i.e., they were
  simply discarded from queries), which can be seen as a lower bound
  method in this comparison,
\item transliteration was applied to every {\it katakana\/} word,
  instead of consulting the base word dictionary.
\end{itemize}
Both cases were combined into the CWT
Section~\ref{subsec:quantitative}. Note that in the latter case, when
a {\it katakana\/} word is included in a compound word,
transliteration candidates of the word are disambiguated through the
compound word translation method, and thus noisy candidates are
potentially discarded.  It should also be noted that in the case where
a compound word consists of solely {\it katakana\/} words (e.g., {\it
deeta-mainingu\/}~(data mining)), our method automatically segments it
into base words, by transliterating all the possible substrings.

Table~\ref{tab:avg_pre_kana} shows the average precision values,
averaged over the 21 queries, for those above cases.  By comparing
Tables~\ref{tab:avg_pre} and \ref{tab:avg_pre_kana}, one can see that
the performance was considerably degraded when we disregard every {\it
katakana\/} word, and that even when we applied transliteration to
every katakana word, the performance was greater than that of CWT
and was quite comparable to that of TRL. Among the 20 {\it
katakana\/} base words, only ``{\it eejento\/}~(agent)'' was
incorrectly transliterated into ``eagent,'' which was due to an
insufficient volume of the transliteration dictionary.

\begin{table}[htbp]
  \def\baselinestretch{1}
  \begin{center}
    \caption{Query-by-query comparison between CWT and TRL}
    \medskip
    \leavevmode
    \footnotesize
    \begin{tabular}{cllll} \hline\hline
      & & \multicolumn{3}{c}{Change in Average Precision (CWT
      $\rightarrow$ TRL)} \\ \cline{3-5}
      ID & Japanese (Translation in TRL) & {\hfill\centering Standard
      TF\hfill} & {\hfill\centering Logarithmic
      TF\hfill} & {\hfill\centering SMART\hfill} \\ \hline
      0008 & {\it deeta-$\underline{mainingu}$\/} (data
      $\underline{\mbox{mining}}$) & 0.0018 $\rightarrow$ 0.0942 &
      0.0299 $\rightarrow$ 0.3363 & 0.3156 $\rightarrow$ 0.7295 \\
      0012 & {\it deeta-$\underline{mainingu}$\/} (data
      $\underline{\mbox{mining}}$) & 0.0018 $\rightarrow$ 0.1229 &
      0.0003 $\rightarrow$ 0.1683 & 0.0000 $\rightarrow$ 0.0853 \\
      0015 & {\it $\underline{corokeishon}$\/}
      ($\underline{\mbox{collocation}}$) & 0.0054 $\rightarrow$ 0.0084
      & 0.0389 $\rightarrow$ 0.0485 & 0.0193 $\rightarrow$
      0.3114 \smallskip \\ \hline
    \end{tabular}
    \label{tab:avgpre_qbq_trl}
  \end{center}
\end{table}

\begin{table}[htbp]
  \def\baselinestretch{1}
  \begin{center}
    \caption{Non-interpolated average precision values,
    averaged over the 21 queries, for the evaluation of
    transliteration}
    \medskip
    \leavevmode
    \small
    \begin{tabular}{lccccc} \hline\hline
      & & \multicolumn{3}{c}{Retrieval Method} \\ \cline{3-5}
      & \# of Terms & Standard TF & Logarithmic TF & SMART \\ \hline
      discard every {\it katakana\/} word & 2.8 & 0.1519 & 0.1840 &
      0.1873 \\
      transliterate every {\it katakana\/} word & 4.0 & 0.2354 & 0.2786 &
      0.3024 \\ \hline
    \end{tabular}
    \label{tab:avg_pre_kana}
  \end{center}
\end{table}

Finally, we discuss the effect of additional dictionaries, i.e., the
general and abbreviation dictionaries. The NACSIS query set contains
the general word ``{\it shimbun kiji\/}~(newspaper article)'' and
abbreviation ``LFG~(lexical functional grammar)'' unlisted in our
technical base word dictionary. The abbreviation dictionary lists the
correct translation for ``LFG.''  On the other hand, our general
dictionary, which consists solely of single words, does not list the
correct translation for ``{\it shimbun-kiji\/}.'' Instead, the English
word ``story'' was listed as the translation, which would be used in a
particular context. Table~\ref{tab:avgpre_qbq_additional}, where basic
notation is the same as Table~\ref{tab:avgpre_qbq_cwt}, compares
average precision values with/without these translations. From this
table we cannot see any improvement with the additional
dictionaries. However, when the correct translation was provided as in
0023 with ``newspaper article,'' the performance was improved
disregarding the retrieval method. In addition, since we found only
two cases where additional dictionaries could be applied, this issue
needs to be further explored using more test queries.

\begin{table}[htbp]
  \def\baselinestretch{1}
  \begin{center}
    \caption{Query-by-query comparison for the general and
    abbreviation dictionaries}
    \medskip
    \leavevmode
    \footnotesize
    \begin{tabular}{cllll} \hline\hline
      & & \multicolumn{3}{c}{Change in Average Precision} \\ \cline{3-5}
      ID & Japanese (Translation) & {\hfill\centering Standard
      TF\hfill} & {\hfill\centering Logarithmic
      TF\hfill} & {\hfill\centering SMART\hfill} \\ \hline
      0023 & {\it shimbun-kiji\/} (story) & 0.0003
      $\rightarrow$ 0.0000* & 0.0000 $\rightarrow$ 0.0000 & 0.0000
      $\rightarrow$ 0.0000 \\
      0023 & {\it shimbun-kiji\/} (newspaper article) & 0.0003
      $\rightarrow$ 0.0200 & 0.0000 $\rightarrow$ 0.0858 & 0.0000
      $\rightarrow$ 0.1800 \\
      0025 & LFG (lexical functional grammar) & 0.8000 $\rightarrow$
      0.5410* & 0.8000 $\rightarrow$ 0.6879* & 0.9452 $\rightarrow$
      0.8617* \\ \hline
    \end{tabular}
    \label{tab:avgpre_qbq_additional}
  \end{center}
\end{table}

\section{Conclusion}
\label{sec:conclusion}

Reflecting the rapid growth in utilization of machine readable
multilingual texts in the 1990s, cross-language information retrieval
(CLIR), which was initiated in the 1960s, has variously been explored
in order to facilitate retrieving information across languages.  For
this purpose, a number of CLIR systems have been developed in
information retrieval, natural language processing and artificial
intelligence research.

In this paper, we proposed a Japanese/English bidirectional CLIR
system targeting technical documents, in that translation of technical
terms is a crucial task. Since our research methodology must be
contextualized in terms of past research literature, we surveyed
existing CLIR systems, and classified them into three approaches: (a)
translating queries into the document language, (b) translating
documents into the query language, and (c) representing both queries
and documents in a language-independent space. Among these approaches,
we found that the first one, namely the query translation approach, is
relatively inexpensive to implement. Therefore, following this
approach, we combined query translation and monolingual retrieval
modules.

However, a naive query translation method relying on existing
bilingual dictionaries does not guarantee sufficient system
performance, because new technical terms are progressively created by
combining existing base words or by the Japanese {\it katakana\/}
phonograms. To counter this problem, we proposed compound word
translation and transliteration methods, and integrated them within
one framework. Our methods involve the dictionary production and
probabilistic resolution of translation/transliteration ambiguity,
both of which are fully automated. To produce the dictionary used for
the compound word translation, we extracted base word translations
from the EDR technical terminology dictionary. On the other hand, we
corresponded English and Japanese {\it katakana\/} words on a
character basis, to produce the transliteration dictionary. For the
disambiguation, we used word frequency statistics extracted from the
document collection. We also produced a dictionary for abbreviated
English technical terms, to enhance the translation.

From a scientific point of view, we investigated the performance of
our CLIR system by way of the standardized IR evaluation method. For
this purpose, we used the NACSIS test collection, which consists of
Japanese queries and Japanese/English technical abstracts, and carried
out Japanese-English CLIR evaluation. Our evaluation results showed
that each individual method proposed, i.e., compound word translation
and transliteration, improved on the baseline performance, and when
used together the improvement was even greater, resulting in a
performance comparable with Japanese-Japanese monolingual IR. We also
showed that the enhancement of the retrieval module improved on our
system performance, independently from the enhancement of the query
translation module.

Future work will include improvement of each component in our system,
and the effective presentation of retrieved documents using
sophisticated summarization techniques.

\section*{Acknowledgments}

The authors would like to thank Noriko Kando (National Institute of
Informatics, Japan) for her support with the NACSIS collection.

\bibliographystyle{acl}

\begin{thebibliography}{}

\bibitem[\protect\citename{AAAI}1997]{aaai-spring-sympo-97}
AAAI.
\newblock 1997.
\newblock {\em Electronic Working Notes of the AAAI Spring Symposium on
  Cross-Language Text and Speech Retrieval}.
\newblock {\tt http://www.clis.umd.edu/dlrg/filter/sss/papers/}.

\bibitem[\protect\citename{ACM}1996-1998]{sigir-96-98}
ACM SIGIR.
\newblock 1996-1998.
\newblock {\em Proceedings of the Annual International ACM SIGIR Conference on
  Research and Development in Information Retrieval}.

\bibitem[\protect\citename{Aone \bgroup et al.\egroup }1997]{aone:anlp-97}
Chinatsu Aone, Nicholas Charocopos, and James Gorlinsky.
\newblock 1997.
\newblock An intelligent multilingual information browsing and retrieval system
  using information extraction.
\newblock In {\em Proceedings of the 5th Conference on Applied Natural Language
  Processing}, pages 332--339.

\bibitem[\protect\citename{Ballesteros and Croft}1997]{ballesteros:sigir-97}
Lisa Ballesteros and W.~Bruce Croft.
\newblock 1997.
\newblock Phrasal translation and query expansion techniques for cross-language
  information retrieval.
\newblock In {\em Proceedings of the 20th Annual International ACM SIGIR
  Conference on Research and Development in Information Retrieval}, pages
  84--91.

\bibitem[\protect\citename{Ballesteros and Croft}1998]{ballesteros:sigir-98}
Lisa Ballesteros and W.~Bruce Croft.
\newblock 1998.
\newblock Resolving ambiguity for cross-language retrieval.
\newblock In {\em Proceedings of the 21st Annual International ACM SIGIR
  Conference on Research and Development in Information Retrieval}, pages
  64--71.

\bibitem[\protect\citename{Brown \bgroup et al.\egroup }1993]{brown:cl-93}
Peter~F. Brown, Stephen A.~Della Pietra, Vincent J.~Della Pietra, and Robert~L.
  Mercer.
\newblock 1993.
\newblock The mathematics of statistical machine translation: Parameter
  estimation.
\newblock {\em Computational Linguistics}, 19(2):263--311.

\bibitem[\protect\citename{Carbonell \bgroup et al.\egroup
  }1997]{carbonell:ijcai-97}
Jaime~G. Carbonell, Yiming Yang, Robert~E. Frederking, Ralf~D. Brown, Yibing
  Geng, and Danny Lee.
\newblock 1997.
\newblock Translingual information retrieval: A comparative evaluation.
\newblock In {\em Proceedings of the 15th International Joint Conference on
  Artificial Intelligence}, pages 708--714.

\bibitem[\protect\citename{Chen \bgroup et al.\egroup
  }1998]{chen:coling-acl-98}
Hsin-Hsi Chen, Sheng-Jie Huang, Yung-Wei Ding, and Shih-Chung Tsai.
\newblock 1998.
\newblock Proper name translation in cross-language information retrieval.
\newblock In {\em Proceedings of the 36th Annual Meeting of the Association for
  Computational Linguistics and the 17th International Conference on
  Computational Linguistics}, pages 232--236.

\bibitem[\protect\citename{Chen \bgroup et al.\egroup }1999]{chen:acl-99}
Hsin-Hsi Chen, Guo-Wei Bian, and Wen-Cheng Lin.
\newblock 1999.
\newblock Resolving translation ambiguity and target polysemy in cross-language
  information retrieval.
\newblock In {\em Proceedings of the 37th Annual Meeting of the Association for
  Computational Linguistics}, pages 215--222.

\bibitem[\protect\citename{Church and Mercer}1993]{church:cl-93}
Kenneth~W. Church and Robert~L. Mercer.
\newblock 1993.
\newblock Introduction to the special issue on computational linguistics using
  large corpora.
\newblock {\em Computational Linguistics}, 19(1):1--24.

\bibitem[\protect\citename{Davis and Ogden}1997]{davis:sigir-97}
Mark~W. Davis and William~C. Ogden.
\newblock 1997.
\newblock {QUILT}: Implementing a large-scale cross-language text retrieval
  system.
\newblock In {\em Proceedings of the 20th Annual International ACM SIGIR
  Conference on Research and Development in Information Retrieval}, pages
  92--98.

\bibitem[\protect\citename{Deerwester \bgroup et al.\egroup
  }1990]{deerwester:jasis-90}
Scott Deerwester, Susan~T. Dumais, George~W. Furnas, Thomas~K. Landauer, and
  Richard Harshman.
\newblock 1990.
\newblock Indexing by latent semantic analysis.
\newblock {\em Journal of the American Society for Information Science},
  41(6):391--407.

\bibitem[\protect\citename{Dijkstra}1959]{dijkstra:nm-59}
Edsgar~W. Dijkstra.
\newblock 1959.
\newblock A note on two problems in connexion with graphs.
\newblock {\em Numerische Mathematik}, 1:269--271.

\bibitem[\protect\citename{Dorr and Oard}1998]{dorr:lrec-98}
Bonnie~J. Dorr and Douglas~W. Oard.
\newblock 1998.
\newblock Evaluating resources for query translation in cross-language
  information retrieval.
\newblock In {\em Proceedings of the 1st International Conference on Language
  Resources and Evaluation}, pages 759--764.

\bibitem[\protect\citename{Dumais \bgroup et al.\egroup
  }1996]{dumais:sigir-ws-96}
Susan~T. Dumais, Thomas~K. Landauer, and Michael~L. Littman.
\newblock 1996.
\newblock Automatic cross-linguistic information retrieval using latent
  semantic indexing.
\newblock In {\em ACM SIGIR Workshop on Cross-Linguistic Information
  Retrieval}.

\bibitem[\protect\citename{Fellbaum}1998]{fellbaum:wordnet-98}
Christiane Fellbaum, editor.
\newblock 1998.
\newblock {\em {WordNet}: An Electronic Lexical Database}.
\newblock MIT Press.

\bibitem[\protect\citename{Ferber}1989]{ferber:89}
Gene Ferber.
\newblock 1989.
\newblock {\em {English-Japanese}, {Japanese-English} Dictionary of Computer
  and Data-Processing Terms}.
\newblock MIT Press.

\bibitem[\protect\citename{Fung \bgroup et al.\egroup }1999]{fung:acl-99}
Pascale Fung, Liu Xiaohu, and Cheung~Chi Shun.
\newblock 1999.
\newblock Mixed language query disambiguation.
\newblock In {\em Proceedings of the 37th Annual Meeting of the Association for
  Computational Linguistics}, pages 333--340.

\bibitem[\protect\citename{Fung}1995]{fung:acl-95}
Pascale Fung.
\newblock 1995.
\newblock A pattern matching method for finding noun and proper noun
  translations from noisy parallel corpora.
\newblock In {\em Proceedings of the 33rd Annual Meeting of the Association for
  Computational Linguistics}, pages 236--243.

\bibitem[\protect\citename{Gachot \bgroup et al.\egroup
  }1996]{gachot:sigir-ws-96}
Denis~A. Gachot, Elke Lange, and Jin Yang.
\newblock 1996.
\newblock The {SYSTRAN} {NLP} browser: An application of machine translation
  technology in multilingual information retrieval.
\newblock In {\em ACM SIGIR Workshop on Cross-Linguistic Information
  Retrieval}.

\bibitem[\protect\citename{Gilarranz \bgroup et al.\egroup
  }1997]{gilarranz:aaai-spring-sympo-97}
Julio Gilarranz, Julio Gonzalo, and Felisa Verdejo.
\newblock 1997.
\newblock An approach to conceptual text retrieval using the {EuroWordNet}
  multilingual semantic database.
\newblock In {\em Electronic Working Notes of the AAAI Spring Symposium on
  Cross-Language Text and Speech Retrieval}.

\bibitem[\protect\citename{Gonzalo \bgroup et al.\egroup
  }1998]{gonzalo:chum-98}
Julio Gonzalo, Felisa Verdejo, Carol Peters, and Nicoletta Calzolari.
\newblock 1998.
\newblock Applying {EuroWordNet} to cross-language text retrieval.
\newblock {\em Computers and the Humanities}, 32:185--207.

\bibitem[\protect\citename{Hull and Grefenstette}1996]{hull:sigir-96}
David~A. Hull and Gregory Grefenstette.
\newblock 1996.
\newblock Querying across languages: A dictionary-based approach to
  multilingual information retrieval.
\newblock In {\em Proceedings of the 19th Annual International ACM SIGIR
  Conference on Research and Development in Information Retrieval}, pages
  49--57.

\bibitem[\protect\citename{Hull}1993]{hull:sigir-93}
David Hull.
\newblock 1993.
\newblock Using statistical testing in the evaluation of retrieval experiments.
\newblock In {\em Proceedings of the 16th Annual International ACM SIGIR
  Conference on Research and Development in Information Retrieval}, pages
  329--338.

\bibitem[\protect\citename{Hull}1997]{hull:aaai-spring-sympo-97}
David~A. Hull.
\newblock 1997.
\newblock Using structured queries for disambiguation in cross-language
  information retrieval.
\newblock In {\em Electronic Working Notes of the AAAI Spring Symposium on
  Cross-Language Text and Speech Retrieval}.

\bibitem[\protect\citename{{Japan Electronic Dictionary Research
  Institute}}1995a]{edr-bilindic:95}
{Japan Electronic Dictionary Research Institute}.
\newblock 1995a.
\newblock Bilingual dictionary.
\newblock (In Japanese).

\bibitem[\protect\citename{{Japan Electronic Dictionary Research
  Institute}}1995b]{edr-techdic:95}
{Japan Electronic Dictionary Research Institute}.
\newblock 1995b.
\newblock Technical terminology dictionary (information processing).
\newblock (In Japanese).

\bibitem[\protect\citename{Kaji and Aizono}1996]{kaji:coling-96}
Hiroyuki Kaji and Toshiko Aizono.
\newblock 1996.
\newblock Extracting word correspondences from bilingual corpora based on word
  co-occurrence information.
\newblock In {\em Proceedings of the 16th International Conference on
  Computational Linguistics}, pages 23--28.

\bibitem[\protect\citename{Kando \bgroup et al.\egroup }1999]{kando:sigir-99}
Noriko Kando, Kazuko Kuriyama, and Toshihiko Nozue.
\newblock 1999.
\newblock {NACSIS} test collection workshop ({NTCIR-1}).
\newblock In {\em Proceedings of the 22nd Annual International ACM SIGIR
  Conference on Research and Development in Information Retrieval}, pages
  299--300.

\bibitem[\protect\citename{Keen}1992]{keen:ipm-92}
E.~Michael Keen.
\newblock 1992.
\newblock Presenting results of experimental retrieval comparisons.
\newblock {\em Information Processing \& Management}, 28(4):491--502.

\bibitem[\protect\citename{Knight and Graehl}1998]{knight:cl-98}
Kevin Knight and Jonathan Graehl.
\newblock 1998.
\newblock Machine transliteration.
\newblock {\em Computational Linguistics}, 24(4):599--612.

\bibitem[\protect\citename{Kobayashi \bgroup et al.\egroup
  }1994]{kobayashi:coling-94}
Yoshiyuki Kobayashi, Takenobu Tokunaga, and Hozumi Tanaka.
\newblock 1994.
\newblock Analysis of {Japanese} compound nouns using collocational
  information.
\newblock In {\em Proceedings of the 15th International Conference on
  Computational Linguistics}, pages 865--869.

\bibitem[\protect\citename{Kwon \bgroup et al.\egroup }1998]{kwon:cpol-98}
Oh-Woog Kwon, Insu Kang, Jong-Hyeok Lee, and Geunbae Lee.
\newblock 1998.
\newblock Conceptual cross-language text retrieval based on document
  translation using {Japanese}-to-{Korean} {MT} system.
\newblock {\em International Journal of Computer Processing of Oriental
  Languages}, 12(1):1--16.

\bibitem[\protect\citename{Lee and Choi}1997]{lee:iral-97}
Jae~Sung Lee and Key-Sun Choi.
\newblock 1997.
\newblock A statistical method to generate various foreign word
  transliterations in multilingual information retrieval system.
\newblock In {\em Proceedings of the 2nd International Workshop on Information
  Retrieval with Asian Languages}, pages 123--128.

\bibitem[\protect\citename{Mani and Bloedorn}1998]{mani:aaai-iaai-98}
Inderjeet Mani and Eric Bloedorn.
\newblock 1998.
\newblock Machine learning of generic and user-focused summarization.
\newblock In {\em Proceedings of AAAI/IAAI-98}, pages 821--826.

\bibitem[\protect\citename{Matsumoto \bgroup et al.\egroup
  }1997]{matsumoto:chasen-97}
Yuji Matsumoto, Akira Kitauchi, Tatsuo Yamashita, Osamu Imaichi, and Tomoaki
  Imamura.
\newblock 1997.
\newblock {Japanese} morphological analysis system {ChaSen} manual.
\newblock Technical Report NAIST-IS-TR97007, NAIST.
\newblock (In Japanese).

\bibitem[\protect\citename{McCarley}1999]{mccarley:acl-99}
J.~Scott McCarley.
\newblock 1999.
\newblock Should we translate the documents or the queries in cross-language
  information retrieval?
\newblock In {\em Proceedings of the 37th Annual Meeting of the Association for
  Computational Linguistics}, pages 208--214.

\bibitem[\protect\citename{Mongar}1969]{mongar:tis-69}
P.E. Mongar.
\newblock 1969.
\newblock International co-operation in abstracting services for road
  engineering.
\newblock {\em The Information Scientist}, 3:51--62.

\bibitem[\protect\citename{{Nichigai Associates}}1996]{nichigai_compdic:96}
{Nichigai Associates}.
\newblock 1996.
\newblock {English-Japanese} computer terminology dictionary.
\newblock (In Japanese).

\bibitem[\protect\citename{Nie \bgroup et al.\egroup }1999]{nie:sigir-99}
Jian-Yun Nie, Michel Simard, Pierre Isabelle, and Richard Durand.
\newblock 1999.
\newblock Cross-language information retrieval based on parallel texts and
  automatic mining of parallel texts from the {Web}.
\newblock In {\em Proceedings of the 22nd Annual International ACM SIGIR
  Conference on Research and Development in Information Retrieval}, pages
  74--81.

\bibitem[\protect\citename{NIST}1992-1998]{trec-92-98}
{National Institute of Standards \& Technology}.
\newblock 1992--1998.
\newblock {\em Proceedings of the Text REtrieval Conferences}.
\newblock {\tt http://trec.nist.gov/pubs.html}.

\bibitem[\protect\citename{Oard and Resnik}1999]{oard:ipm-99}
Douglas~W. Oard and Philip Resnik.
\newblock 1999.
\newblock Support for interactive document selection in cross-language
  information retrieval.
\newblock {\em Information Processing \& Management}, 35(3):363--379.

\bibitem[\protect\citename{Oard}1998]{oard:amta-98}
Douglas~W. Oard.
\newblock 1998.
\newblock A comparative study of query and document translation for
  cross-language information retrieval.
\newblock In {\em Proceedings of the 3rd Conference of the Association for
  Machine Translation in the Americas}, pages 472--483.

\bibitem[\protect\citename{Okumura \bgroup et al.\egroup
  }1998]{okumura:lrec-tlim-ws-98}
Akitoshi Okumura, Kai Ishikawa, and Kenji Satoh.
\newblock 1998.
\newblock Translingual information retrieval by a bilingual dictionary and
  comparable corpus.
\newblock In {\em The 1st International Conference on Language Resources and
  Evaluation, Workshop on Translingual Information Management: Current Levels
  and Future Abilities}.

\bibitem[\protect\citename{Pirkola}1998]{pirkola:sigir-98}
Ari Pirkola.
\newblock 1998.
\newblock The effects of query structure and dictionary setups in
  dictionary-based cross-language information retrieval.
\newblock In {\em Proceedings of the 21st Annual International ACM SIGIR
  Conference on Research and Development in Information Retrieval}, pages
  55--63.

\bibitem[\protect\citename{Sakai \bgroup et al.\egroup }1999]{sakai:tipsj-99}
Tetsuya Sakai, Masahiro Kajiura, Kazuo Sumita, Gareth Jones, and Nigel Collier.
\newblock 1999.
\newblock A study on {English}-{Japanese}/{Japanese}-{English} cross-language
  information retrieval using machine translation.
\newblock {\em Transactions of Information Processing Society of Japan},
  40(11):4075--4086.
\newblock (In Japanese).

\bibitem[\protect\citename{Salton and Buckley}1988]{salton:ipm-88}
Gerard Salton and Christopher Buckley.
\newblock 1988.
\newblock Term-weighting approaches in automatic text retrieval.
\newblock {\em Information Processing \& Management}, 24(5):513--523.

\bibitem[\protect\citename{Salton and McGill}1983]{salton:83}
Gerard Salton and Michael~J. McGill.
\newblock 1983.
\newblock {\em Introduction to Modern Information Retrieval}.
\newblock McGraw-Hill.

\bibitem[\protect\citename{Salton}1970]{salton:jasis-70}
Gerard Salton.
\newblock 1970.
\newblock Automatic processing of foreign language documents.
\newblock {\em Journal of the American Society for Information Science},
  21(3):187--194.

\bibitem[\protect\citename{Salton}1971]{salton:71}
Gerard Salton.
\newblock 1971.
\newblock {\em The {SMART} Retrieval System: Experiments in Automatic Document
  Processing}.
\newblock Prentice-Hall.

\bibitem[\protect\citename{Salton}1972]{salton:techrep-72}
Gerard Salton.
\newblock 1972.
\newblock Experiments in multi-lingual information retrieval.
\newblock Technical Report TR 72-154, Computer Science Department, Cornell
  University.

\bibitem[\protect\citename{Sch\"{a}uble and Sheridan}1997]{schauble:trec-97}
Peter Sch\"{a}uble and P\'{a}raic Sheridan.
\newblock 1997.
\newblock Cross-language information retrieval ({CLIR}) track overview.
\newblock In {\em {\it The 6th Text Retrieval Conference}}.

\bibitem[\protect\citename{Sheridan and Ballerini}1996]{sheridan:sigir-96}
P\'{a}raic Sheridan and Jean~Paul Ballerini.
\newblock 1996.
\newblock Experiments in multilingual information retrieval using the {SPIDER}
  system.
\newblock In {\em Proceedings of the 19th Annual International ACM SIGIR
  Conference on Research and Development in Information Retrieval}, pages
  58--65.

\bibitem[\protect\citename{Smadja \bgroup et al.\egroup }1996]{smadja:cl-96}
Frank Smadja, Kathleen~R. McKeown, and Vasileios Hatzivassiloglou.
\newblock 1996.
\newblock Translating collocations for bilingual lexicons: A statistical
  approach.
\newblock {\em Computational Linguistics}, 22(1):1--38.

\bibitem[\protect\citename{Suzuki \bgroup et al.\egroup
  }1998]{suzuki:signl-98-7}
Masami Suzuki, Naomi Inoue, and Kazuo Hashimoto.
\newblock 1998.
\newblock Effect on displaying translated major keywords of contents as
  browsing support in cross-language information retrieval.
\newblock {\em Information Processing Society of Japan SIGNL Notes},
  98(63):99--106.
\newblock (In Japanese).

\bibitem[\protect\citename{Suzuki \bgroup et al.\egroup }1999]{suzuki:nlp-99}
Masami Suzuki, Naomi Inoue, and Kazuo Hashimoto.
\newblock 1999.
\newblock Effects of partial translation for users' document selection in
  cross-language information retrieval.
\newblock In {\em Proceedings of The 5th Annual Meeting of The Association for
  Natural Language Processing}, pages 371--374.
\newblock (In Japanese).

\bibitem[\protect\citename{Tombros and Sanderson}1998]{tombros:sigir-98}
Anastasios Tombros and Mark Sanderson.
\newblock 1998.
\newblock Advantages of query biased summaries in information retrieval.
\newblock In {\em Proceedings of the 21st Annual International ACM SIGIR
  Conference on Research and Development in Information Retrieval}, pages
  2--10.

\bibitem[\protect\citename{Tsuji and Kageura}1997]{tsuji:nlprs-97}
Keita Tsuji and Kyo Kageura.
\newblock 1997.
\newblock An {HMM}-based method for segmenting {Japanese} terms and keywords
  based on domain-specific bilingual corpora.
\newblock In {\em Proceedings of the 4th Natural Language Processing Pacific
  Rim Symposium}, pages 557--560.

\bibitem[\protect\citename{Voorhees}1998]{voorhees:sigir-98}
Ellen~M. Voorhees.
\newblock 1998.
\newblock Variations in relevance judgments and the measurement of retrieval
  effectiveness.
\newblock In {\em Proceedings of the 21st Annual International ACM SIGIR
  Conference on Research and Development in Information Retrieval}, pages
  315--323.

\bibitem[\protect\citename{Vossen}1998]{vossen:chum-98}
Piek Vossen.
\newblock 1998.
\newblock Introduction to {EuroWordNet}.
\newblock {\em Computers and the Humanities}, 32:73--89.

\bibitem[\protect\citename{Wong \bgroup et al.\egroup }1985]{wong:sigir-85}
S.K.M. Wong, W.~Siarko, and P.C.N. Wong.
\newblock 1985.
\newblock Generalized vector space model in information retrieval.
\newblock In {\em Proceedings of the 8th Annual International ACM SIGIR
  Conference on Research and Development in Information Retrieval}, pages
  18--25.

\bibitem[\protect\citename{Xu and Croft}1996]{xu:sigir-96}
Jinxi Xu and W.~Bruce Croft.
\newblock 1996.
\newblock Query expansion using local and global document analysis.
\newblock In {\em Proceedings of the 19th Annual International ACM SIGIR
  Conference on Research and Development in Information Retrieval}, pages
  4--11.

\bibitem[\protect\citename{Yamabana \bgroup et al.\egroup
  }1996]{yamabana:sigir-ws-96}
Kiyoshi Yamabana, Kazunori Muraki, Shinichi Doi, and Shin'ichiro Kamei.
\newblock 1996.
\newblock A language conversion front-end for cross-linguistic information
  retrieval.
\newblock In {\em ACM SIGIR Workshop on Cross-Linguistic Information
  Retrieval}.

\bibitem[\protect\citename{Zobel and Moffat}1998]{zobel:sigir-forum-98}
Justin Zobel and Alistair Moffat.
\newblock 1998.
\newblock Exploring the similarity space.
\newblock {\em ACM SIGIR FORUM}, 32(1):18--34.

\end{thebibliography}

\end{document}